\documentclass{article} 
\usepackage{iclr2025_conference}

\usepackage{amsmath,amsfonts,bm}

\def\eqref#1{equation~\ref{#1}}

\def\1{\bm{1}}

\DeclareMathAlphabet{\mathsfit}{\encodingdefault}{\sfdefault}{m}{sl}
\SetMathAlphabet{\mathsfit}{bold}{\encodingdefault}{\sfdefault}{bx}{n}

\usepackage{hyperref}
\usepackage{url}
\usepackage{graphicx}
\usepackage{amsmath}
\usepackage{subcaption}
\usepackage{float}
\usepackage{tcolorbox}
\usepackage{xcolor}
\usepackage{textcomp}
\usepackage{placeins}

\title{BRAID: Bounded Reasoning for Autonomous Inference and Decisions}

\author{Armağan Amcalar \thanks{Work performed while at OpenServ Labs} \\
Chief Technology Officer\\
OpenServ Labs \\
\texttt{armagan@openserv.ai} \\
\And
Eyup Cinar \thanks{Work performed while at OpenServ Labs as AI Research Partner} \\
Computer Engineering Department \\
Eskisehir Osmangazi University \\
\texttt{eyup.cinar@ogu.edu.tr}
}

\iclrfinalcopy 
\begin{document}

\maketitle

\begin{abstract}
Large Language Models (LLMs) exhibit nonlinear relationships between performance, cost, and token usage. This paper presents a quantitative study on structured prompting using BRAID (Bounded Reasoning for Autonomous Inference and Decisions) across multiple GPT model tiers, evaluated on the AdvancedIF, GSM-Hard, and the SCALE MultiChallenge benchmark datasets. BRAID introduces a bounded reasoning framework using Mermaid-based instruction graphs that enable models to reason structurally rather than through unbounded natural-language token expansion. We show that structured machine-readable prompts substantially increase reasoning accuracy and cost efficiency for agents in production systems. The findings establish BRAID as an effective and scalable technique for optimizing inference efficiency in autonomous agent systems. All datasets and detailed result logs are available at \url{https://benchmark.openserv.ai}.
\end{abstract}

\section{Introduction}

Large Language Models (LLMs) have achieved remarkable success on many NLP tasks, especially as their scale reaches hundreds of billions of parameters. Although each newer model is marketed with advanced reasoning capabilities, their cost efficiency remains a bottleneck for many companies and practitioners.

Early research work demonstrated that large language models can be prompted to perform new tasks without gradient updates by providing task descriptions or examples in natural language. An influential example is GPT-3 (175B parameters), which demonstrated that in-context learning via few-shot prompting can achieve strong performance on diverse NLP tasks using only text demonstrations instead of fine-tuning \cite{brown2020language}. In this standard prompting paradigm, a model is given either zero examples (zero-shot) or a handful of input-output examples (few-shot) before a query, and the model is expected to infer the pattern and produce the correct output. GPT-3’s few-shot results showed that scaling up the model size produces impressive zero- and few-shot reasoning abilities in translation, question answering, and even simple arithmetic without task-specific training. However these in-context learning strategies showed \textit{shallow reasoning} with limitations on complex reasoning tasks. This spurred the development of new prompting strategies, especially elicited step-by-step reasoning from LLMs.

\subsection {Chain-of-Thought Prompting and Unstructured Reasoning}

Chain-of-Thought (CoT) prompting is a landmark prompting strategy that elicits intermediate reasoning steps before the final answer. In CoT prompting, the few-shot exemplars are augmented with explicit step by-step solutions (“thoughts”) instead of just input–output pairs. \cite{wei2022chain} showed that even a handful of such worked examples can dramatically improve performance on arithmetic, commonsense, and symbolic reasoning tasks.

 After CoT’s introduction, researchers discovered LLMs can produce reasoning steps even without example demonstrations. \cite{kojima2022large} found that simply appending a prompt like “Let’s think step by step” to the query triggers many language models to generate a coherent chain of thought in a zero-shot setting. This Zero-Shot CoT approach revealed that LLMs are “decent zero-shot reasoners” when encouraged to articulate multi-step solutions, often dramatically improving accuracy over direct answers (e.g. boosting GPT-3’s math word problem accuracy from 10\%  to 40\% on GSM8K). These findings underscored that even without explicit training, large models harbor latent reasoning capabilities that can be unlocked by an \textit{appropriate prompt}.

  Despite CoT’s success, its free-form reasoning traces can sometimes be incorrect or suboptimal. One mitigation is to sample multiple distinct chains of thought and aggregate their answers – the Self Consistency decoding strategy \cite{wang2022self}. Rather than relying on a single CoT, self-consistency samples a diverse set of reasoning paths and then takes a majority vote or consensus on the final answer. However, this approach requires multiple input and output model query turns and can be operationally more expensive with respect to single run model with a prompting strategy. As pointed out by \cite{sprague2024tocot}, entire new paradigms, possibly involving \textit{external symbolic tools or computations}, will be needed to extend reasoning improvements to the full range of LLM applications.
  
  \subsection {Structured and Enhanced Prompting Approaches}

Researchers have developed more \textit{structured prompting strategies} to further improve or generalize chain-of-thought reasoning. These methods introduce additional guidance, formatting, or intermediate steps in prompts to tackle complex problems more reliably. Decompositional prompting is one of them proposed by \cite{zhou2022least} as \textit{``least-to-most prompting''} technique. In this technique a complex problem is broken into a sequence of simpler sub-problems  which the model solves one by one. The prompt first asks the model to derive a small intermediate question, then uses the answer to that sub-problem to inform the next step, and so on.  By chaining these incremental resolutions, the model can handle problems more difficult than those seen in the prompt examples. Zhou et al. reported significant accuracy gains over standard CoT by these methodology.  This approach highlights how \textit{explicit decomposition} in prompts enables better generalization to hard tasks by reducing them to manageable pieces. 

Plan-and-Solve  prompting \cite{wang2023plan} uses a two-phase prompt to avoid missing steps in zero-shot reasoning. First, the model is prompted to “devise a plan” – a high-level outline of steps or sub-tasks needed to solve the problem. Next, the model is prompted to “solve” each sub-task according to that plan. By explicitly structuring the reasoning process into a planning stage and an execution stage, plan-and-solve addresses errors where a vanilla CoT might skip necessary steps or jump to conclusions. Experiments showed zero-shot Plan-and-Solve prompting can outperform the basic Zero-Shot CoT across a variety of tasks, and even approach the performance of few-shot CoT.

Another frontier is automating the prompt design for reasoning tasks. Universal Self-Adaptive Prompting (USP) \cite{wan2023universal} is an approach that learns to construct effective prompts for arbitrary tasks in a zero-shot way.  The technique uses a small unlabeled corpus and the LLM itself to generate candidate reasoning examples: it first classifies a given task into one of a few broad types (question answering, translation, etc.), then selects representative queries and model-generated solutions as pseudo-demonstrations to include in a prompt.  In essence, the model is helping to “prompt itself” by producing its own chain-of-thought exemplars, eliminating manual prompt engineering. This automated CoT prompting was shown to significantly outperform naive zero-shot prompts and even match or exceed few-shot performance on many benchmarks. Wan et al. report that USP, when applied to PaLM and PaLM-2 models, achieved results \textit{“often comparable to or even superior to few-shot baselines”} across 40+ tasks, including reasoning-intensive benchmarks.  By generalizing in-context learning to the zero-shot regime, USP represents a structured prompting framework that adapts to each new task with minimal human intervention.  

Classic Chain-of-Thought (CoT) reasoning inherently increases token length and introduces linguistic noise—many of these tokens carry low semantic density but still incur cost. This can potentially degrade the signal-to-noise ratio, particularly for high-end reasoning models like GPT-5 with high reasoning effort, which often generate verbose reasoning traces without proportional accuracy improvement \cite{zhou2024canlanguagemodelsperform}, \cite{lin2024canlargelanguagemodels}. Due to these limitations, the field has moved towards structured prompting techniques. The Bounded Reasoning for Autonomous Inference and Decisions (BRAID) framework is proposed as a structural improvement to classical prompting techniques. BRAID’s core mechanism is the replacement of natural-language reasoning with bounded, symbolic structures. By constraining the reasoning path to deterministic logical flows expressed in Mermaid diagrams, BRAID aims to yield more compact reasoning with significantly higher token efficiency. Rather than letting the model “think aloud,” BRAID constrains reasoning paths to deterministic logical flows. This structural shift not only mitigates the token-cost problem but also reduces “reasoning drift” where the model’s internal monologue becomes verbose or off-topic.

\section{BRAID: A Novel Reasoning Methodology}
The BRAID framework is a novel approach to structured prompting that fundamentally alters the nature of the LLM's internal reasoning process from an unbounded linguistic monologue to a bounded, symbolic plan. BRAID replaces the natural-language CoT trace with a \textbf{bounded, symbolic reasoning structure} expressed in Mermaid diagrams. The core principle is to compress the cognitive process into high-density tokens by encoding the logical flow as a diagram rather than a descriptive text. This structural constraint ensures that the model's reasoning path is deterministic, reducing the possibility of "reasoning drift" (off-topic or repetitive text) that plagues traditional CoT.

\begin{figure}[h]
  \centering
  \includegraphics[width=1\linewidth]{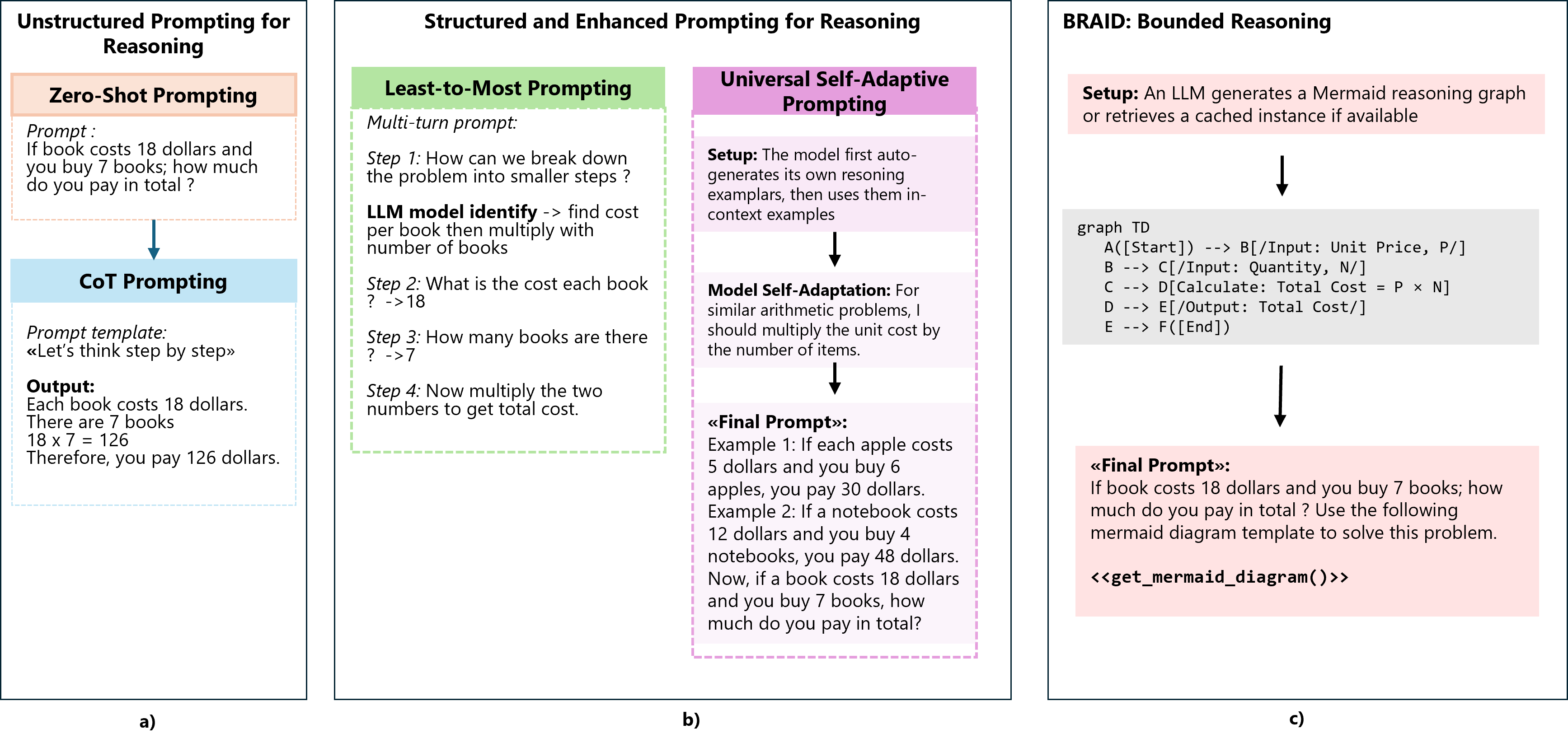}
  \caption{a) Unstructured prompting encourages models to show intermediate steps in natural language before the answer b) Structured and Enhanced prompting techniques explicitly decomposes the problem into simpler sub-problems and solve sequentially c) BRAID replaces the natural-language prompt trace with structured, symbolic reasoning paths expressed in Mermaid diagrams }
  \label{fig:placeholder}
\end{figure}

The key contributions of this work are:
\begin{enumerate}
    \item \textbf{Comparable accuracy gains:} Demonstrating that encoding reasoning steps in structured symbolic form (Mermaid diagrams) with low capacity models leads to same level of accuracy and consistency with brand new, significantly more expensive and state-of-the-art models.
    \item \textbf{Quantitative Economic Analysis:} Revealing a crucial finding for LLM cost reduction and showing how BRAID can enable LLM cost reduction for end users. 
    \item \textbf{Efficiency Gains:} Introducing performance-per-dollar (PPD) metric for quantifying LLM efficiency gains,  achieving major   leap in optimal configurations, particularly on smaller or cheaper models, which is crucial for deploying generative agents.
\end{enumerate}

The study explores how BRAID affects both accuracy and cost efficiency across various state-of-the-art OpenAI's GPT model tiers using three well-known benchmark datasets such as GSM-Hard which  originates from the PaLM-r paper by \cite{gao2022palr}, SCALE MultiChallenge by \cite{sirdeshmukh2025multichallenge}  and AdvancedIF benchmarks by \cite{advancedif2025}. Beyond empirical performance, we analyze the economic implications by mapping the relationship between model pricing and performance-per-dollar across all configurations.

\section{Related Work}
As indicated in the first section, prior work on prompt engineering and reasoning optimization has focused on natural-language techniques
such as chain-of-thought prompting, self-consistency, and reflection loops. While these approaches improve
reasoning, they also increase token usage, leading to higher inference costs.

 The literature suggests several mechanistic rationales for why structured prompting helps. Explicit stepwise instructions e.g. Tree of Thoughts by \cite{yao2023_ToT} scaffold the model’s internal generation toward intermediate reasoning tokens and expose latent multi‑step chains that yield better final answers.  \cite{dutta2024thinkstepby} conducted a mechanistic interpretability analysis of the LLaMA-2 model and found that LLMs internally organize CoT reasoning into distinct pathways. In  the transformer's layers: earlier layers remain biased toward retrieving world knowledge (the pretraining prior), while later layers shift focus on the in-context reasoning provided by the CoT prompt. Thus, structured prompting helps to activate later layers for logical combination. 

 The literature also suggests that decomposing a complex problem into a sequence of smaller sub problems also helps with improving reasoning. As studied by \cite{wang2023planandsolve} in Plan-and-Solve approach, the technique first asks the model to outline a plan (a sequence of subtask instructions) and then executes that plan stepwise. This was shown to reduce errors like missing steps or calculation mistakes common in free-form CoT, yielding at least 5 \% accuracy improvements in zero-shot mathematical reasoning and commonsense QA over CoT.

 Program-of-Thought (PoT) or Chain-of-Symbol approaches by \cite{chen2023programthoughtspromptingdisentangling} replace or supplement natural language steps with executable code or symbolic representations. For example, the technique lets the model generate Python code for calculations (and then run it), offloading the “compute” part to a reliable executor. This reduces the model’s burden in tasks like math by avoiding manual arithmetic errors – indeed PoT improved math problem accuracy by ~12\% over plain CoT. These methods show that structured prompts that mimic formal reasoning (code, equations) can enhance LLM reasoning, presumably by leveraging the model’s training on such formats and by providing unambiguous intermediate steps.

For code-generation tasks, Structured Chain-of-Thought (SCoT) prompting by \cite{li2023structuredchainofthoughtpromptingcode} uses pseudo-code structures (like loops, branches) as the “thoughts” to guide code writing. This was inspired by the idea that a coder first sketches program logic before coding. SCoT improved accuracy in code generation by up to 13.8\% over plain natural language reasoning, indicating that mirroring the semantic structure of the task in the prompt leads to better-organized solutions.
 
 These recent studies have explored structured reasoning representations, but none have quantitatively linked token economics with bounded reasoning architectures. Our proposed technique BRAID fills that gap by reframing structured prompting as a measurable performance–cost function, rather than a qualitative enhancement.

\section{Experimental Setup}
We evaluated 272 SCALE MultiChallenge reasoning questions, 100 GSM-Hard and 100 AdvancedIF questions across combinations of GPT models acting as prompt generators and solvers. Each test followed a two-stage protocol: 1. Prompt Generation: Each model generated a BRAID (Mermaid) diagram encoding the logical reasoning process for the question. 2. Prompt Solving: Another model (same or different tier) used the BRAID as a system message and produced the final answer.
 This produced a matrix of model-pair combinations (e.g., Medium → Nano, Nano Minimal → Mini Medium). Token usage and API cost were tracked per run. Performance was measured as accuracy (\%), and efficiency as performance per dollar, normalized against \texttt{gpt-5-medium} baseline = 1.0. All questions from the datasets were randomly selected except SCALE MultiChallange dataset.

Regarding the prompting setup, for the control conditions, we employed a strict zero-shot prompting protocol, deliberately omitting explicit Chain-of-Thought triggers (e.g., `Let's think step by step'). This methodological choice was dictated by the architectural nature of the GPT-5 model family. With the exception of \texttt{gpt-5.1-none} and \texttt{gpt-4o}, the models evaluated in this study possess intrinsic latent reasoning capabilities (`thinking models') that automatically engage during inference. Consequently, forcing an external Chain-of-Thought layer would introduce redundancy and conflate the model’s native reasoning with the prompt's instructions. By relying on a raw zero-shot input for the baseline, we isolate the specific impact of the BRAID structure against the model’s native, unguided performance.

Across all three datasets, accuracy was evaluated using an external LLM, with GPT-5.2 (medium reasoning effort) serving as the adjudicator. We deliberately chose this protocol over strict equality testing—even for the deterministic GSM-Hard benchmark—to avoid enforcing rigid structured output schemas (e.g., JSON). We observed that forcing the model to segregate its computational process into specific schema fields disrupts its natural generation trajectory, often degrading performance. By utilizing an LLM judge, we permitted the solver models to produce responses in their native, unconstrained free-form style. This ensures that metrics reflect true reasoning capabilities rather than adherence to syntactic constraints. We utilized automated LLM-generated reasoning graphs as a scalable alternative to manually optimizing graphs for each unique question in the diverse dataset. However, in a production setting, our approach is designed to leverage predefined, manually handcrafted reasoning plans that can be cached and reused repeatedly.

To ensure the rigor of our evaluation specifically on the GSM-Hard benchmark, we implemented a Numerical Masking Protocol to address ``answer leakage.'' Since generating a logic graph for math often necessitates calculating intermediate values, the generator model frequently transcribes these directly into the node labels. To prevent the Solver from merely retrieving these pre-calculated solutions, a post-processing step parses the Mermaid diagram and replaces all numerical literals with a neutral placeholder (i.e., \texttt{\_}). This ensures the artifact conveys only the \textit{logical topology} while withholding the computational state. Because the model operates in the unconstrained format described above, it utilizes the requisite token volume to perform the arithmetic operations necessitated by these masked nodes. The Judge LLM then verifies the correctness of this final numerical derivation against the ground truth.

 \subsection{Cost and Performance-per-Dollar (PPD) Metric}
 Two cost (\textit{C}) perspectives were analyzed:  
\noindent
where:
\begin{itemize}
    \item $C_{\text{BRAID}}$ denotes the cost of generating structured reasoning or a BRAID-style prompt with Mermaid diagrams. This step typically includes the reasoning synthesis, planning, or decomposition overhead.
    \item $C_{\text{inference}}$ represents the cost of executing the model on a given prompt or reasoning chain to produce the final output.
\end{itemize}

\vspace{1em}
For deployed agents or large-scale systems where BRAID prompts are either written by hand or generated once and reused many times, it is relevant to consider the \textbf{Solving-Only Cost}:

\begin{equation}
C_{\text{solve-only}} = C_{\text{inference}}
\label{eq:1}
\end{equation}

\noindent
This formulation excludes BRAID generation, which is negligible when prompts are cached or reused millions of times during inference.

\vspace{1em}
In cases where BRAID generation is performed once but reused across $N$ queries, the amortized total cost per query can be expressed as:

\begin{equation}
C_{\text{amortized}} = \frac{C_{\text{BRAID}}}{N} + C_{\text{inference}}
\label{eq:2}
\end{equation}

\noindent
This amortized form captures the realistic long-term cost distribution for agents that reuse structured prompts, balancing one-time reasoning setup costs with per-query inference expenses.

Each cost was computed as the sum of token count multiplied by its corresponding unit price, aggregated per question and normalized to United States Dollars (USD). Formally, for a given model:

\begin{equation}
C_{\text{model}} = \sum_{i=1}^{Q} \left( T_{\text{in}, i} \cdot p_{\text{in}} + T_{\text{out}, i} \cdot p_{\text{out}} \right)
\label{eq:3}
\end{equation}

\noindent
where:
\begin{itemize}
    \item $Q$ denotes the number of evaluation questions or tasks,
    \item $T_{\text{in}, i}$ and $T_{\text{out}, i}$ represent the number of input and output tokens, respectively, for the $i^{\text{th}}$ query,
    \item $p_{\text{in}}$ and $p_{\text{out}}$ correspond to the unit token prices (USD per token) for input and output processing.
\end{itemize}

\vspace{1em}

To enable a fair comparison across models with differing pricing and accuracy levels, the \textbf{Performance-per-Dollar (PPD)} metric was defined as follows:

\begin{equation}
    \text{PPD} = 
    \frac{
        \displaystyle\frac{\text{Accuracy}}{\text{Cost}}
    }{
        \displaystyle\frac{\text{Accuracy}_\mathrm{GPT5\text{-}Medium}}{\text{Cost}_\mathrm{GPT5\text{-}Medium}}
    }
    \label{eq:ppd_definition}
\end{equation}
\noindent

Here, the normalization is performed relative to the GPT-5-Medium model, ensuring that $\text{PPD} = 1$ for the reference system. Values greater than 1 indicate superior cost-efficiency, while values below 1 indicate less efficient performance per unit cost. This allows direct comparison across models independent of raw accuracy or absolute cost.

\clearpage

\section{Results}
We evaluated the BRAID framework across three distinct benchmarks: GSM-Hard, SCALE MultiChallenge, and AdvancedIF. For each benchmark, we compared the accuracy, cost, and Performance-per-Dollar (PPD) of models using classic unstructured prompting versus BRAID structured prompting. 

\subsection{Accuracy Analysis}
The implementation of BRAID yielded significant accuracy improvements across all model tiers and datasets, most notably enabling smaller models to perform on par with or surpass larger models using classic prompting  and the models' inherent reasoning efforts.

\begin{figure}[!htbp]
    \centering
    
    \begin{subfigure}[b]{1.0\linewidth}
        \centering
        \includegraphics[width=\linewidth]{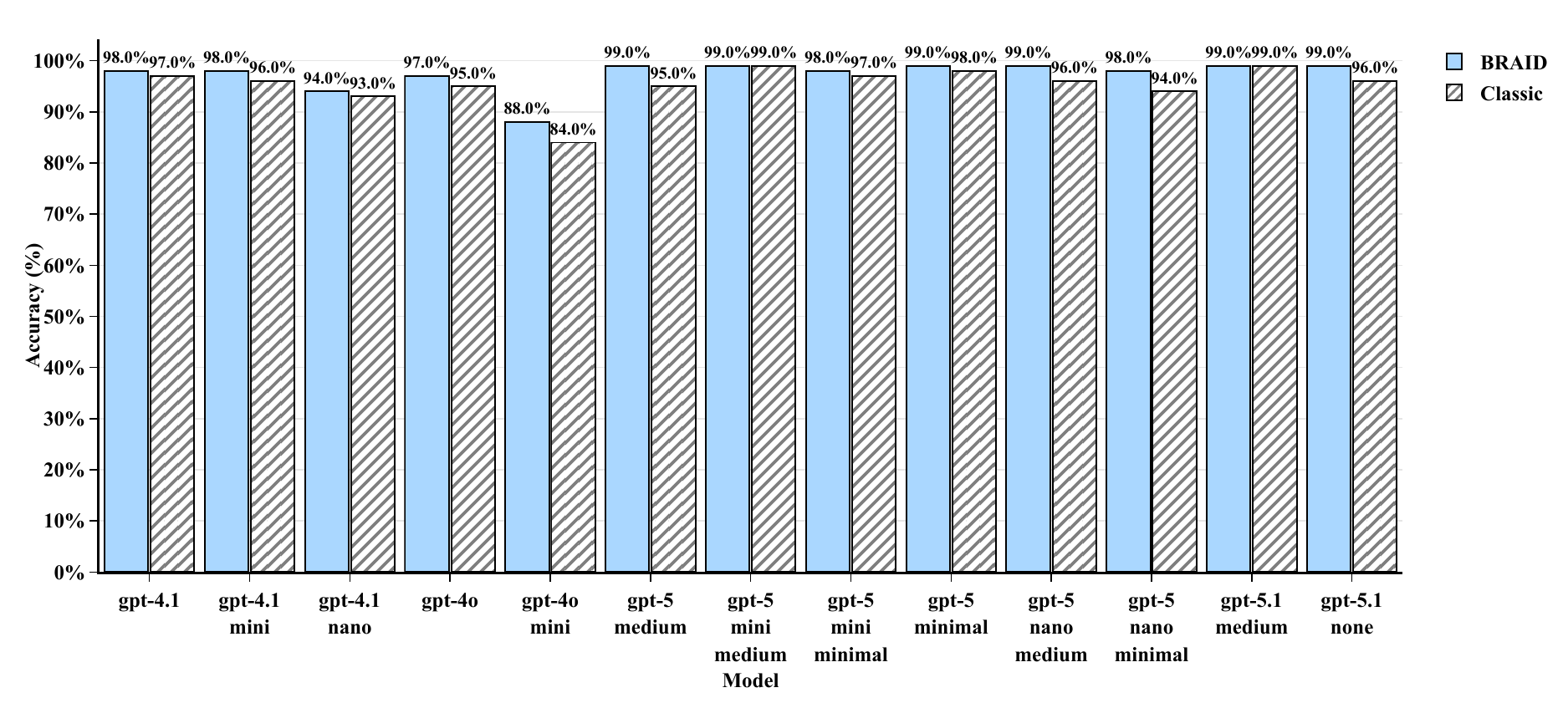}
        \caption{GSM-Hard Dataset}
        \label{fig:gsm_acc}
    \end{subfigure}
    
    \vspace{0.5em} 
    
    \begin{subfigure}[b]{1.0\linewidth}
        \centering
        \includegraphics[width=\linewidth]{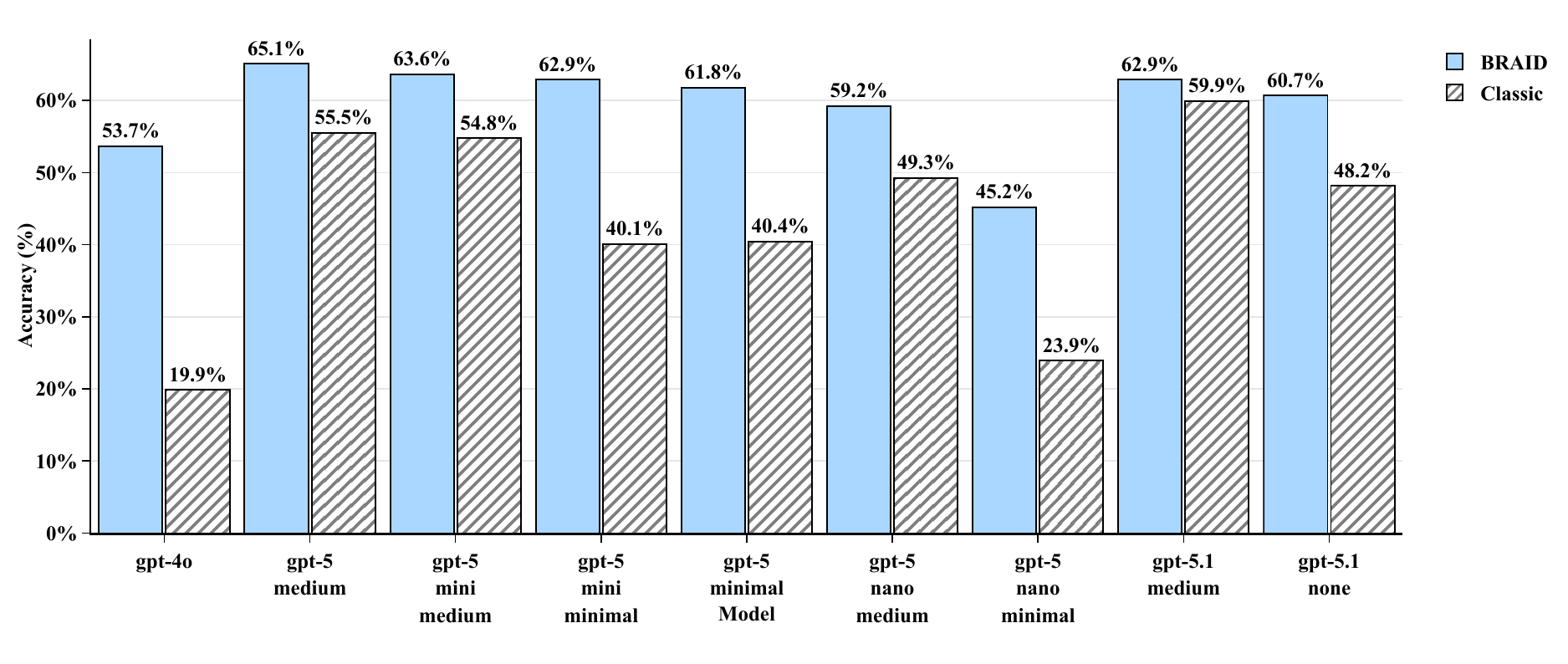}
        \caption{SCALE MultiChallenge Dataset}
        \label{fig:scale_acc}
    \end{subfigure}
    
    \vspace{0.5em} 
    
    \begin{subfigure}[b]{1.0\linewidth}
        \centering
        \includegraphics[width=\linewidth]{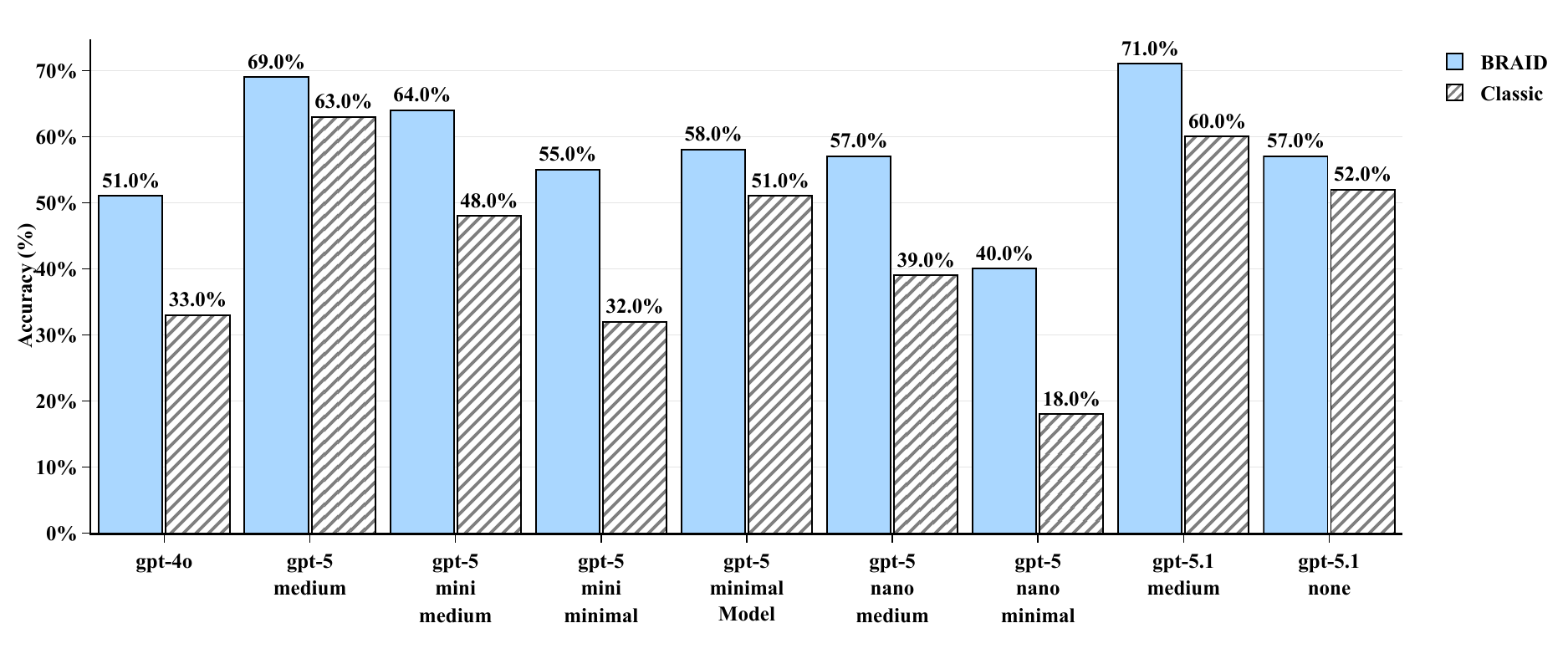}
        \caption{AdvancedIF Dataset}
        \label{fig:if_acc}
    \end{subfigure}
    
    \caption{\textbf{Comparative Reasoning Accuracy of BRAID vs Classic Prompting Best for each Solving Model:} BRAID (blue) can enable smaller models to match or exceed the performance of larger models using Classic (hatched) across (a) GSM-Hard, (b) SCALE MultiChallenge, and (c) AdvancedIF instruction following benchmarks.}
    \label{fig:accuracy_comparison}
\end{figure}

Figure~\ref{fig:accuracy_comparison} presents the best for each solving accuracy results for the three datasets. Best for each solving model means that for each solving model, the reported BRAID accuracy represents the maximum performance achieved across all tested generator combinations, isolating the model's peak execution capability. As shown in Fig. \ref{fig:gsm_acc}, the GSM-Hard benchmark is effectively saturated by the evaluated models. Unlike the procedural datasets where smaller models struggled significantly without structure, the intrinsic mathematical capabilities of the GPT-5 and GPT-4 families resulted in high "Classic" baselines (typically $>90\%$). However, BRAID maintained a consistent edge, pushing performance from "excellent" to "near-perfect." For instance, \texttt{gpt-5-nano-minimal} improved from 94.0\% (Classic) to \textbf{98.0\%} (BRAID), and \texttt{gpt-5-medium} improved from 95.0\% to \textbf{99.0\%}. This demonstrates that even when models possess strong latent reasoning, the BRAID framework's bounded structure can bridge the final gap to precision, achieving results that are indistinguishable from the upper bound of the benchmark.

As shown in Fig. \ref{fig:scale_acc} for SCALE MultiChallenge dataset which involves complex multi-step reasoning, BRAID showed the largest relative gains. \texttt{gpt-4o} saw a massive performance leap from 19.9\% with Classic prompting to 53.7\% with BRAID. Similarly, the cost-efficient \texttt{gpt-5-nano-minimal} improved from 23.9\% to 45.2\%, outperforming the Classic implementation of the much larger \texttt{gpt-5-minimal} (40.4\%). A significant finding was that \texttt{gpt-5-medium} with BRAID was able to outperform \texttt{gpt-5-medium}.

In the text-based interactions of AdvancedIF, BRAID maintained a consistent lead as shown in Fig. \ref{fig:if_acc}. \texttt{gpt-5.1-medium} reached 71.0\% accuracy with BRAID compared to 60.0\% with Classic prompting. Notably, the gap was most pronounced in smaller models; \texttt{gpt-5-nano-minimal} more than doubled its accuracy from 18.0\% to 40.0\%.

\subsection{Cost Analysis}
\begin{figure}[!h]
    \centering

    \begin{subfigure}[b]{1.0\linewidth} 
        \centering
        \includegraphics[width=\linewidth]{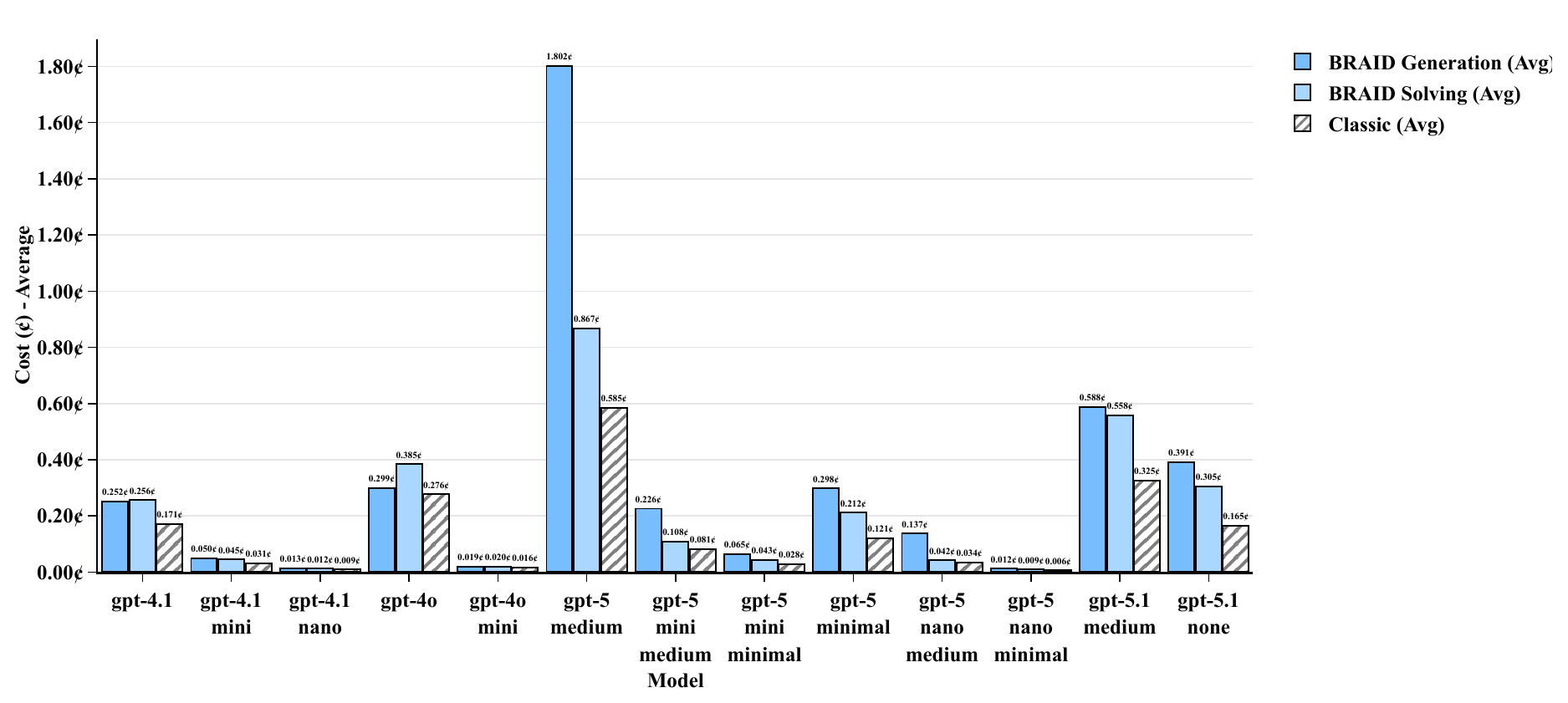}
        \caption{GSM-Hard Dataset}
        \label{fig:gsm_cost}
    \end{subfigure}

    \begin{subfigure}[b]{1.0\linewidth}
        \centering
        \includegraphics[width=\linewidth]{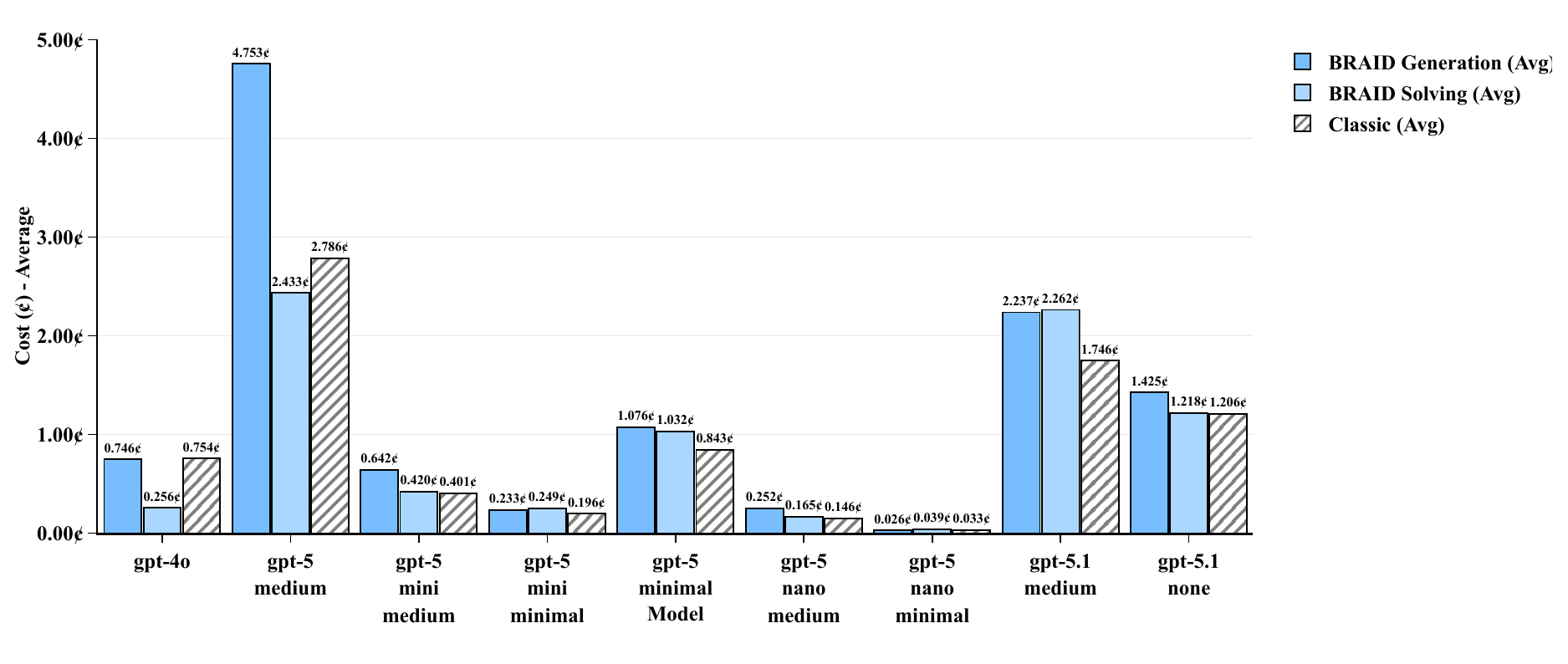}
        \caption{SCALE MultiChallenge Dataset}
        \label{fig:scale_cost}
    \end{subfigure}
    
    \begin{subfigure}[b]{1.0\linewidth}
        \centering
        \includegraphics[width=\linewidth]{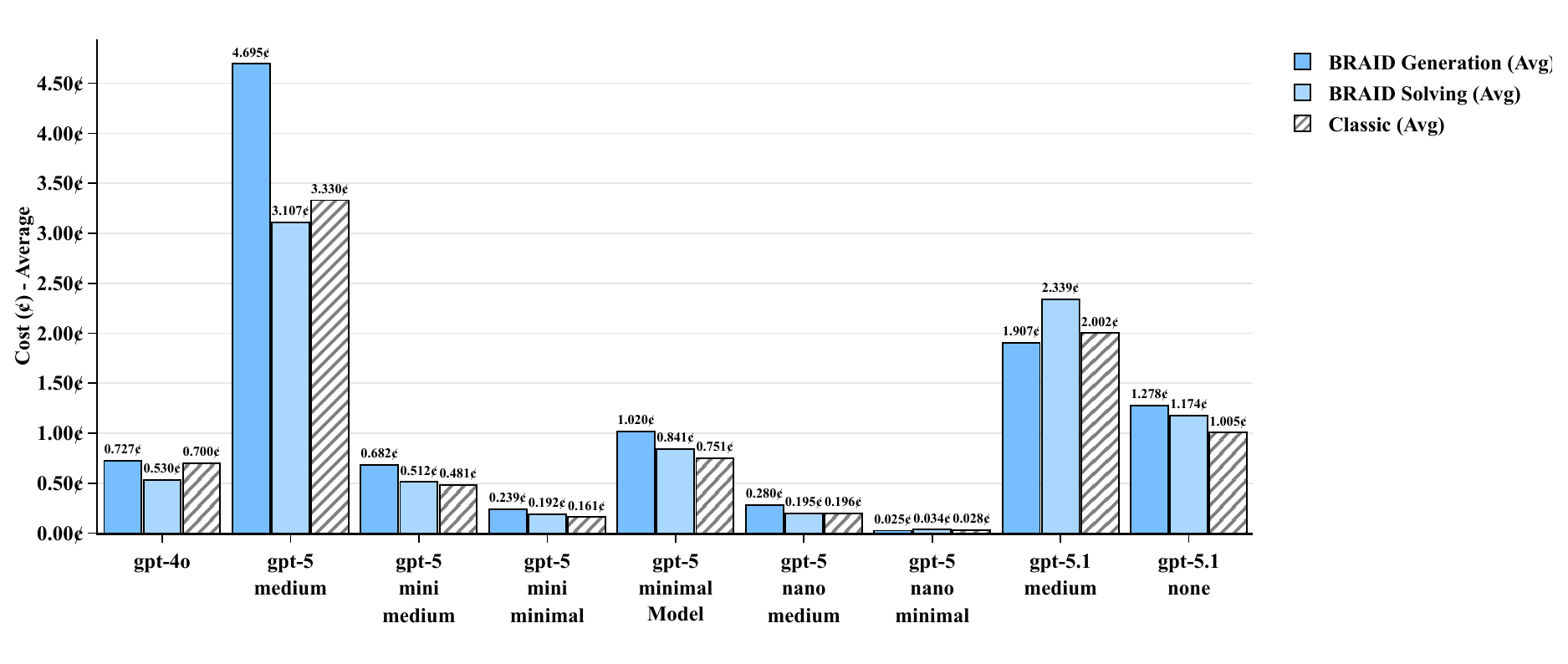}
        \caption{AdvancedIF Dataset}
        \label{fig:if_cost}
    \end{subfigure}
    
   \caption{\textbf{Average Cost Breakdown per Response (in US cents):} Results contrast the inference costs of BRAID Generation and Solving phases against the Classic prompting baseline across various model architectures. Notably, the solving costs (light blue) for smaller models are significantly lower than the baseline, demonstrating a major economic advantage for agentic workflows that leverage cached Mermaid reasoning graphs.}
    \label{fig:cost_analysis}
\end{figure}

We analyzed the cost footprint of our approach by initially calculating the average cost per response (in US cents, \textcent) across three distinct datasets: GSM-Hard, AdvancedIF, and SCALE MultiChallenge. Fig. \ref{fig:cost_analysis} illustrates the cost breakdown across three dimensions: BRAID Generation, BRAID Solving, and the Classic baseline.

The cost of generating Mermaid diagram structures varied considerably across models. However, as noted in eq. \ref{eq:2}, this cost is amortized in agentic workflows. Cost comparisons reveal that the marginal gap between BRAID’s solving phase and the Classic baseline is minimal. Moreover, in certain cases, BRAID's solving costs fall below the Classic baseline.

\FloatBarrier
\subsection{Performance-per-Dollar (PPD) Efficiency}
The PPD metric in eq. \ref{eq:ppd_definition} highlights the economic significance of the BRAID framework. By decoupling reasoning (generation) from execution (solving), we identified a "Golden Quadrant" of efficiency: using high-intelligence models for generation and low-cost models for solving.

As presented in Table \ref{table:GSMPPD1}, the efficiency gains for GSM-Hard confirm the viability of this split-architecture approach. A standout configuration pairs \texttt{gpt-4.1} for generation with \texttt{gpt-5-nano-minimal} for solving, achieving a PPD of 74.06 relative to the \texttt{gpt-5-medium} baseline. This result indicates a greater than 74-fold improvement in cost-effectiveness while achieving a better performance (96\% accuracy vs. the baseline's 95\%). This empirically validates that high-precision mathematical reasoning does not strictly require monolithic high-cost models when the reasoning topology is decoupled from calculation.

\vspace{1cm}
\begin{table}[!h]
\centering
\caption{Accuracy \% and Performance per Dollar (PPD) for GSM-Hard Dataset}
\label{table:GSMPPD1}
{\scriptsize
\setlength{\arraycolsep}{3pt}
\renewcommand{\arraystretch}{1.15}
\resizebox{\textwidth}{!}{$
\begin{array}{l|ccccccccccccc} 
\multicolumn{1}{c|}{\mathbf{Gen} \rightarrow \mathbf{Solve}} & \texttt{gpt-4.1} & \begin{array}{c}\texttt{gpt-4.1}\\\texttt{mini}\end{array} & \begin{array}{c}\texttt{gpt-4.1}\\\texttt{nano}\end{array} & \texttt{gpt-4o} & \begin{array}{c}\texttt{gpt-4o}\\\texttt{mini}\end{array} & \begin{array}{c}\texttt{gpt-5}\\\texttt{medium}\end{array} & \begin{array}{c}\texttt{gpt-5}\\\texttt{mini}\\\texttt{medium}\end{array} & \begin{array}{c}\texttt{gpt-5}\\\texttt{mini}\\\texttt{minimal}\end{array} & \begin{array}{c}\texttt{gpt-5}\\\texttt{minimal}\end{array} & \begin{array}{c}\texttt{gpt-5}\\\texttt{nano}\\\texttt{medium}\end{array} & \begin{array}{c}\texttt{gpt-5}\\\texttt{nano}\\\texttt{minimal}\end{array} & \begin{array}{c}\texttt{gpt-5.1}\\\texttt{medium}\end{array} & \begin{array}{c}\texttt{gpt-5.1}\\\texttt{none}\end{array} \\ \hline
\texttt{ gpt-4.1} & \begin{array}{c}98.0\%\\2.67\end{array} & \begin{array}{c}97.0\%\\14.86\end{array} & \begin{array}{c}93.0\%\\52.20\end{array} & \begin{array}{c}97.0\%\\1.68\end{array} & \begin{array}{c}82.0\%\\27.84\end{array} & \begin{array}{c}98.0\%\\0.77\end{array} & \begin{array}{c}98.0\%\\5.77\end{array} & \begin{array}{c}98.0\%\\15.62\end{array} & \begin{array}{c}98.0\%\\3.15\end{array} & \begin{array}{c}98.0\%\\15.30\end{array} & \begin{array}{c}\textbf{96.0}\%\\\textbf{74.06}\end{array} & \begin{array}{c}98.0\%\\1.22\end{array} & \begin{array}{c}97.0\%\\2.26\end{array} \\ \hline
\begin{array}{l}\texttt{gpt-4.1}\\\texttt{mini}\end{array} & \begin{array}{c}98.0\%\\2.63\end{array} & \begin{array}{c}98.0\%\\15.18\end{array} & \begin{array}{c}91.0\%\\50.74\end{array} & \begin{array}{c}93.0\%\\1.68\end{array} & \begin{array}{c}84.0\%\\28.32\end{array} & \begin{array}{c}98.0\%\\0.77\end{array} & \begin{array}{c}98.0\%\\6.05\end{array} & \begin{array}{c}97.0\%\\15.92\end{array} & \begin{array}{c}98.0\%\\3.14\end{array} & \begin{array}{c}98.0\%\\15.36\end{array} & \begin{array}{c}95.0\%\\69.71\end{array} & \begin{array}{c}98.0\%\\1.22\end{array} & \begin{array}{c}98.0\%\\2.21\end{array} \\ \hline
\begin{array}{l}\texttt{gpt-4.1}\\\texttt{nano}\end{array} & \begin{array}{c}96.0\%\\2.51\end{array} & \begin{array}{c}95.0\%\\14.66\end{array} & \begin{array}{c}88.0\%\\49.33\end{array} & \begin{array}{c}94.0\%\\1.65\end{array} & \begin{array}{c}81.0\%\\26.61\end{array} & \begin{array}{c}97.0\%\\0.68\end{array} & \begin{array}{c}97.0\%\\5.52\end{array} & \begin{array}{c}96.0\%\\15.11\end{array} & \begin{array}{c}96.0\%\\3.06\end{array} & \begin{array}{c}97.0\%\\13.54\end{array} & \begin{array}{c}94.0\%\\72.71\end{array} & \begin{array}{c}98.0\%\\1.16\end{array} & \begin{array}{c}97.0\%\\2.25\end{array} \\ \hline
\texttt{ gpt-4o} & \begin{array}{c}98.0\%\\2.64\end{array} & \begin{array}{c}96.0\%\\14.78\end{array} & \begin{array}{c}91.0\%\\51.09\end{array} & \begin{array}{c}97.0\%\\1.65\end{array} & \begin{array}{c}82.0\%\\27.53\end{array} & \begin{array}{c}98.0\%\\0.80\end{array} & \begin{array}{c}98.0\%\\6.15\end{array} & \begin{array}{c}96.0\%\\15.41\end{array} & \begin{array}{c}97.0\%\\3.29\end{array} & \begin{array}{c}98.0\%\\14.73\end{array} & \begin{array}{c}93.0\%\\67.86\end{array} & \begin{array}{c}98.0\%\\1.25\end{array} & \begin{array}{c}98.0\%\\2.17\end{array} \\ \hline
\begin{array}{l}\texttt{gpt-4o}\\\texttt{mini}\end{array} & \begin{array}{c}97.0\%\\2.48\end{array} & \begin{array}{c}95.0\%\\14.29\end{array} & \begin{array}{c}91.0\%\\49.78\end{array} & \begin{array}{c}96.0\%\\1.54\end{array} & \begin{array}{c}88.0\%\\28.53\end{array} & \begin{array}{c}99.0\%\\0.69\end{array} & \begin{array}{c}99.0\%\\5.97\end{array} & \begin{array}{c}98.0\%\\15.18\end{array} & \begin{array}{c}98.0\%\\3.01\end{array} & \begin{array}{c}99.0\%\\14.71\end{array} & \begin{array}{c}97.0\%\\69.15\end{array} & \begin{array}{c}99.0\%\\1.18\end{array} & \begin{array}{c}99.0\%\\2.01\end{array} \\ \hline
\begin{array}{l}\texttt{gpt-5}\\\texttt{medium}\end{array} & \begin{array}{c}97.0\%\\2.35\end{array} & \begin{array}{c}97.0\%\\13.04\end{array} & \begin{array}{c}91.0\%\\46.91\end{array} & \begin{array}{c}92.0\%\\1.48\end{array} & \begin{array}{c}81.0\%\\25.55\end{array} & \begin{array}{c}98.0\%\\0.72\end{array} & \begin{array}{c}98.0\%\\5.58\end{array} & \begin{array}{c}98.0\%\\14.87\end{array} & \begin{array}{c}97.0\%\\2.96\end{array} & \begin{array}{c}98.0\%\\14.67\end{array} & \begin{array}{c}96.0\%\\64.56\end{array} & \begin{array}{c}98.0\%\\1.08\end{array} & \begin{array}{c}98.0\%\\1.81\end{array} \\ \hline
\begin{array}{l}\texttt{gpt-5}\\\texttt{mini}\\\texttt{medium}\end{array} & \begin{array}{c}98.0\%\\2.33\end{array} & \begin{array}{c}98.0\%\\13.09\end{array} & \begin{array}{c}94.0\%\\49.23\end{array} & \begin{array}{c}93.0\%\\1.43\end{array} & \begin{array}{c}85.0\%\\25.89\end{array} & \begin{array}{c}99.0\%\\0.71\end{array} & \begin{array}{c}99.0\%\\5.96\end{array} & \begin{array}{c}98.0\%\\14.38\end{array} & \begin{array}{c}99.0\%\\2.89\end{array} & \begin{array}{c}99.0\%\\14.02\end{array} & \begin{array}{c}96.0\%\\63.77\end{array} & \begin{array}{c}99.0\%\\1.07\end{array} & \begin{array}{c}98.0\%\\1.86\end{array} \\ \hline
\begin{array}{l}\texttt{gpt-5}\\\texttt{mini}\\\texttt{minimal}\end{array} & \begin{array}{c}96.0\%\\2.24\end{array} & \begin{array}{c}92.0\%\\12.11\end{array} & \begin{array}{c}90.0\%\\46.41\end{array} & \begin{array}{c}95.0\%\\1.51\end{array} & \begin{array}{c}78.0\%\\24.06\end{array} & \begin{array}{c}96.0\%\\0.66\end{array} & \begin{array}{c}95.0\%\\5.20\end{array} & \begin{array}{c}95.0\%\\13.05\end{array} & \begin{array}{c}95.0\%\\2.57\end{array} & \begin{array}{c}96.0\%\\14.57\end{array} & \begin{array}{c}92.0\%\\65.69\end{array} & \begin{array}{c}96.0\%\\0.98\end{array} & \begin{array}{c}96.0\%\\1.99\end{array} \\ \hline
\begin{array}{l}\texttt{gpt-5}\\\texttt{minimal}\end{array} & \begin{array}{c}96.0\%\\2.25\end{array} & \begin{array}{c}95.0\%\\12.54\end{array} & \begin{array}{c}93.0\%\\47.61\end{array} & \begin{array}{c}97.0\%\\1.56\end{array} & \begin{array}{c}81.0\%\\25.74\end{array} & \begin{array}{c}98.0\%\\0.72\end{array} & \begin{array}{c}98.0\%\\5.37\end{array} & \begin{array}{c}98.0\%\\14.24\end{array} & \begin{array}{c}96.0\%\\2.78\end{array} & \begin{array}{c}98.0\%\\14.47\end{array} & \begin{array}{c}95.0\%\\66.92\end{array} & \begin{array}{c}98.0\%\\1.11\end{array} & \begin{array}{c}98.0\%\\1.92\end{array} \\ \hline
\begin{array}{l}\texttt{gpt-5}\\\texttt{nano}\\\texttt{medium}\end{array} & \begin{array}{c}96.0\%\\2.52\end{array} & \begin{array}{c}96.0\%\\14.04\end{array} & \begin{array}{c}87.0\%\\47.13\end{array} & \begin{array}{c}94.0\%\\1.54\end{array} & \begin{array}{c}85.0\%\\28.50\end{array} & \begin{array}{c}97.0\%\\0.74\end{array} & \begin{array}{c}97.0\%\\5.60\end{array} & \begin{array}{c}95.0\%\\15.70\end{array} & \begin{array}{c}94.0\%\\3.15\end{array} & \begin{array}{c}96.0\%\\13.40\end{array} & \begin{array}{c}92.0\%\\67.03\end{array} & \begin{array}{c}98.0\%\\1.14\end{array} & \begin{array}{c}97.0\%\\2.04\end{array} \\ \hline
\begin{array}{l}\texttt{gpt-5}\\\texttt{nano}\\\texttt{minimal}\end{array} & \begin{array}{c}94.0\%\\2.30\end{array} & \begin{array}{c}95.0\%\\12.53\end{array} & \begin{array}{c}86.0\%\\44.73\end{array} & \begin{array}{c}92.0\%\\1.48\end{array} & \begin{array}{c}83.0\%\\26.55\end{array} & \begin{array}{c}95.0\%\\0.64\end{array} & \begin{array}{c}95.0\%\\5.23\end{array} & \begin{array}{c}93.0\%\\13.11\end{array} & \begin{array}{c}95.0\%\\2.76\end{array} & \begin{array}{c}92.0\%\\13.28\end{array} & \begin{array}{c}90.0\%\\64.48\end{array} & \begin{array}{c}95.0\%\\1.01\end{array} & \begin{array}{c}94.0\%\\1.98\end{array} \\ \hline
\begin{array}{l}\texttt{gpt-5.1}\\\texttt{medium}\end{array} & \begin{array}{c}98.0\%\\1.95\end{array} & \begin{array}{c}97.0\%\\11.34\end{array} & \begin{array}{c}91.0\%\\44.26\end{array} & \begin{array}{c}96.0\%\\1.36\end{array} & \begin{array}{c}82.0\%\\22.76\end{array} & \begin{array}{c}98.0\%\\0.65\end{array} & \begin{array}{c}98.0\%\\5.24\end{array} & \begin{array}{c}98.0\%\\11.44\end{array} & \begin{array}{c}97.0\%\\2.29\end{array} & \begin{array}{c}98.0\%\\13.21\end{array} & \begin{array}{c}98.0\%\\63.33\end{array} & \begin{array}{c}97.0\%\\0.97\end{array} & \begin{array}{c}98.0\%\\1.73\end{array} \\ \hline
\begin{array}{l}\texttt{gpt-5.1}\\\texttt{none}\end{array} & \begin{array}{c}98.0\%\\1.84\end{array} & \begin{array}{c}97.0\%\\10.74\end{array} & \begin{array}{c}90.0\%\\42.30\end{array} & \begin{array}{c}96.0\%\\1.29\end{array} & \begin{array}{c}87.0\%\\23.74\end{array} & \begin{array}{c}98.0\%\\0.55\end{array} & \begin{array}{c}98.0\%\\4.83\end{array} & \begin{array}{c}98.0\%\\10.67\end{array} & \begin{array}{c}97.0\%\\2.13\end{array} & \begin{array}{c}98.0\%\\14.32\end{array} & \begin{array}{c}98.0\%\\61.54\end{array} & \begin{array}{c}97.0\%\\0.80\end{array} & \begin{array}{c}98.0\%\\1.61\end{array} \\ \hline
\end{array}
$}
}
\end{table}

\clearpage
A similar trend is observed in more complex reasoning tasks presented in other datasets such as SCALE MultiChallenge in Table \ref{table:scaleppd} and Table \ref{tab:advncedifppd} for AdvancedIF. As presented in Table \ref{table:scaleppd} on SCALE MultiChallenge, using \texttt{gpt-5-nano-medium} as a solver yielded a PPD value of 30.31 when paired with a capable generator \texttt{gpt-5-medium}, compared to PPDs of ~1.0-2.0 for monolithic large models. This confirms that small models are capable of high-fidelity execution when the reasoning path is strictly bounded. Furthermore, results in Table \ref{tab:advncedifppd} illustrates a clear strategic trade-off: while the \texttt{gpt-5-nano-minimal} solver yields the highest theoretical efficiency (61.69 PPD), shifting to the \texttt{gpt-5-nano-medium} solver recovers nearly all baseline accuracy (57.0\% vs. 63.0\%) while still delivering a robust 16-fold reduction in cost.

\begin{table}[!h]
\centering
\caption{Accuracy \% and Performance per Dollar (PPD) for SCALE MultiChallenge Dataset}
\label{table:scaleppd}
{\scriptsize
\setlength{\arraycolsep}{3pt}
\renewcommand{\arraystretch}{1.15}
\resizebox{\textwidth}{!}{$
\begin{array}{l|lllllllll}
\hline
\mathbf{Gen} \rightarrow \mathbf{Solve} & \texttt{gpt-4o} & \begin{array}[t]{l}\texttt{gpt-5}\\\texttt{medium}\end{array} & \begin{array}[t]{l}\texttt{gpt-5}\\\texttt{mini}\\\texttt{medium}\end{array} & \begin{array}[t]{l}\texttt{gpt-5}\\\texttt{mini}\\\texttt{minimal}\end{array} & \begin{array}[t]{l}\texttt{gpt-5}\\\texttt{minimal}\end{array} & \begin{array}[t]{l}\texttt{gpt-5}\\\texttt{nano}\\\texttt{medium}\end{array} & \begin{array}[t]{l}\texttt{gpt-5}\\\texttt{nano}\\\texttt{minimal}\end{array} & \begin{array}[t]{l}\texttt{gpt-5.1}\\\texttt{medium}\end{array} & \begin{array}[t]{l}\texttt{gpt-5.1}\\\texttt{none}\end{array} \\ \hline
\texttt{ gpt-4o} & \begin{array}[t]{l}16.9\%\\2.42\end{array} & \begin{array}[t]{l}58.5\%\\0.96\end{array} & \begin{array}[t]{l}57.7\%\\6.18\end{array} & \begin{array}[t]{l}43.4\%\\10.31\end{array} & \begin{array}[t]{l}47.4\%\\2.60\end{array} & \begin{array}[t]{l}44.5\%\\13.16\end{array} & \begin{array}[t]{l}27.9\%\\41.82\end{array} & \begin{array}[t]{l}59.2\%\\1.50\end{array} & \begin{array}[t]{l}51.5\%\\1.83\end{array} \\ \hline
\begin{array}[t]{l}\texttt{gpt-5}\\\texttt{medium}\end{array} & \begin{array}[t]{l}53.7\%\\9.76\end{array} & \begin{array}[t]{l}61.0\%\\1.05\end{array} & \begin{array}[t]{l}63.6\%\\6.93\end{array} & \begin{array}[t]{l}62.9\%\\11.74\end{array} & \begin{array}[t]{l}61.8\%\\2.84\end{array} & \begin{array}[t]{l}\textbf{59.2}\%\\\textbf{30.31}\end{array} & \begin{array}[t]{l}45.2\%\\52.84\end{array} & \begin{array}[t]{l}62.9\%\\1.36\end{array} & \begin{array}[t]{l}60.7\%\\1.75\end{array} \\ \hline
\begin{array}[t]{l}\texttt{gpt-5}\\\texttt{mini}\\\texttt{medium}\end{array} & \begin{array}[t]{l}46.0\%\\10.71\end{array} & \begin{array}[t]{l}60.3\%\\0.85\end{array} & \begin{array}[t]{l}51.5\%\\4.65\end{array} & \begin{array}[t]{l}52.9\%\\9.57\end{array} & \begin{array}[t]{l}56.6\%\\2.36\end{array} & \begin{array}[t]{l}52.6\%\\14.27\end{array} & \begin{array}[t]{l}35.7\%\\43.46\end{array} & \begin{array}[t]{l}61.4\%\\1.14\end{array} & \begin{array}[t]{l}55.1\%\\1.50\end{array} \\ \hline
\begin{array}[t]{l}\texttt{gpt-5}\\\texttt{mini}\\\texttt{minimal}\end{array} & \begin{array}[t]{l}40.4\%\\9.15\end{array} & \begin{array}[t]{l}58.8\%\\0.89\end{array} & \begin{array}[t]{l}54.8\%\\5.29\end{array} & \begin{array}[t]{l}40.1\%\\8.21\end{array} & \begin{array}[t]{l}55.5\%\\3.20\end{array} & \begin{array}[t]{l}48.9\%\\12.60\end{array} & \begin{array}[t]{l}39.3\%\\51.38\end{array} & \begin{array}[t]{l}59.6\%\\1.23\end{array} & \begin{array}[t]{l}55.5\%\\1.63\end{array} \\ \hline
\begin{array}[t]{l}\texttt{gpt-5}\\\texttt{minimal}\end{array} & \begin{array}[t]{l}43.8\%\\7.70\end{array} & \begin{array}[t]{l}55.1\%\\0.90\end{array} & \begin{array}[t]{l}53.3\%\\5.40\end{array} & \begin{array}[t]{l}54.0\%\\10.33\end{array} & \begin{array}[t]{l}40.4\%\\1.84\end{array} & \begin{array}[t]{l}49.6\%\\13.92\end{array} & \begin{array}[t]{l}40.8\%\\48.08\end{array} & \begin{array}[t]{l}57.7\%\\1.22\end{array} & \begin{array}[t]{l}58.8\%\\4.10\end{array} \\ \hline
\begin{array}[t]{l}\texttt{gpt-5}\\\texttt{nano}\\\texttt{medium}\end{array} & \begin{array}[t]{l}35.3\%\\10.55\end{array} & \begin{array}[t]{l}56.3\%\\2.85\end{array} & \begin{array}[t]{l}50.7\%\\9.60\end{array} & \begin{array}[t]{l}42.6\%\\8.91\end{array} & \begin{array}[t]{l}46.3\%\\2.25\end{array} & \begin{array}[t]{l}40.1\%\\11.11\end{array} & \begin{array}[t]{l}34.2\%\\47.89\end{array} & \begin{array}[t]{l}55.5\%\\1.19\end{array} & \begin{array}[t]{l}51.1\%\\1.63\end{array} \\ \hline
\begin{array}[t]{l}\texttt{gpt-5}\\\texttt{nano}\\\texttt{minimal}\end{array} & \begin{array}[t]{l}28.3\%\\3.72\end{array} & \begin{array}{l}57.4\%\\0.89\end{array} & \begin{array}[t]{l}51.1\%\\5.09\end{array} & \begin{array}[t]{l}44.5\%\\9.77\end{array} & \begin{array}[t]{l}46.3\%\\2.30\end{array} & \begin{array}[t]{l}41.2\%\\11.59\end{array} & \begin{array}[t]{l}23.9\%\\31.19\end{array} & \begin{array}[t]{l}55.9\%\\1.24\end{array} & \begin{array}[t]{l}49.3\%\\1.62\end{array} \\ \hline
\begin{array}[t]{l}\texttt{gpt-5.1}\\\texttt{medium}\end{array} & \begin{array}[t]{l}51.8\%\\10.71\end{array} & \begin{array}[t]{l}65.1\%\\1.16\end{array} & \begin{array}[t]{l}58.8\%\\6.78\end{array} & \begin{array}[t]{l}56.3\%\\11.43\end{array} & \begin{array}[t]{l}58.5\%\\2.87\end{array} & \begin{array}[t]{l}56.3\%\\16.54\end{array} & \begin{array}[t]{l}44.9\%\\55.54\end{array} & \begin{array}[t]{l}55.1\%\\1.39\end{array} & \begin{array}[t]{l}57.0\%\\4.84\end{array} \\ \hline
\begin{array}[t]{l}\texttt{gpt-5.1}\\\texttt{none}\end{array} & \begin{array}[t]{l}46.3\%\\7.58\end{array} & \begin{array}[t]{l}55.1\%\\1.22\end{array} & \begin{array}[t]{l}54.8\%\\9.83\end{array} & \begin{array}[t]{l}53.3\%\\10.37\end{array} & \begin{array}[t]{l}49.6\%\\2.44\end{array} & \begin{array}[t]{l}49.3\%\\14.26\end{array} & \begin{array}[t]{l}43.8\%\\55.07\end{array} & \begin{array}[t]{l}59.9\%\\1.53\end{array} & \begin{array}[t]{l}48.2\%\\1.55\end{array} \\ \hline
\end{array}
$}
}
\end{table}

\begin{table}[h!]
\centering
\caption{ Accuracy \% and Performance per Dollar (PPD) for AdvancedIF Dataset}
\label{tab:advncedifppd}
{\scriptsize
\setlength{\arraycolsep}{3pt}
\renewcommand{\arraystretch}{1.15}
\resizebox{\textwidth}{!}{$
\begin{array}{l|lllllllll}
\hline
\mathbf{Gen} \rightarrow \mathbf{Solve} & \texttt{gpt-4o} & \begin{array}[t]{l}\texttt{gpt-5}\\\texttt{medium}\end{array} & \begin{array}[t]{l}\texttt{gpt-5}\\\texttt{mini}\\\texttt{medium}\end{array} & \begin{array}[t]{l}\texttt{gpt-5}\\\texttt{mini}\\\texttt{minimal}\end{array} & \begin{array}[t]{l}\texttt{gpt-5}\\\texttt{minimal}\end{array} & \begin{array}[t]{l}\texttt{gpt-5}\\\texttt{nano}\\\texttt{medium}\end{array} & \begin{array}[t]{l}\texttt{gpt-5}\\\texttt{nano}\\\texttt{minimal}\end{array} & \begin{array}[t]{l}\texttt{gpt-5.1}\\\texttt{medium}\end{array} & \begin{array}[t]{l}\texttt{gpt-5.1}\\\texttt{none}\end{array} \\ \hline
\texttt{ gpt-4o} & \begin{array}[t]{l}34.0\%\\2.39\end{array} & \begin{array}[t]{l}69.0\%\\1.05\end{array} & \begin{array}[t]{l}60.0\%\\5.94\end{array} & \begin{array}[t]{l}48.0\%\\15.54\end{array} & \begin{array}[t]{l}45.0\%\\3.02\end{array} & \begin{array}[t]{l}44.0\%\\11.74\end{array} & \begin{array}[t]{l}27.0\%\\51.47\end{array} & \begin{array}[t]{l}61.0\%\\1.40\end{array} & \begin{array}[t]{l}51.0\%\\2.62\end{array} \\ \hline
\begin{array}[t]{l}\texttt{gpt-5}\\\texttt{medium}\end{array} & \begin{array}[t]{l}51.0\%\\2.73\end{array} & \begin{array}[t]{l}62.0\%\\1.04\end{array} & \begin{array}[t]{l}60.0\%\\6.51\end{array} & \begin{array}[t]{l}55.0\%\\15.34\end{array} & \begin{array}[t]{l}58.0\%\\3.53\end{array} & \begin{array}[t]{l}\textbf{57.0}\%\\\textbf{16.23}\end{array} & \begin{array}[t]{l}40.0\%\\61.69\end{array} & \begin{array}[t]{l}70.0\%\\1.75\end{array} & \begin{array}[t]{l}57.0\%\\2.53\end{array} \\ \hline
\begin{array}[t]{l}\texttt{gpt-5}\\\texttt{mini}\\\texttt{medium}\end{array} & \begin{array}[t]{l}38.0\%\\6.22\end{array} & \begin{array}[t]{l}66.0\%\\0.90\end{array} & \begin{array}[t]{l}54.0\%\\5.14\end{array} & \begin{array}[t]{l}39.0\%\\9.88\end{array} & \begin{array}[t]{l}56.0\%\\2.97\end{array} & \begin{array}[t]{l}47.0\%\\11.45\end{array} & \begin{array}[t]{l}31.0\%\\45.39\end{array} & \begin{array}[t]{l}64.0\%\\1.26\end{array} & \begin{array}[t]{l}50.0\%\\2.08\end{array} \\ \hline
\begin{array}[t]{l}\texttt{gpt-5}\\\texttt{mini}\\\texttt{minimal}\end{array} & \begin{array}[t]{l}38.0\%\\3.05\end{array} & \begin{array}[t]{l}66.0\%\\1.04\end{array} & \begin{array}[t]{l}57.0\%\\5.21\end{array} & \begin{array}[t]{l}39.0\%\\9.87\end{array} & \begin{array}[t]{l}45.0\%\\2.33\end{array} & \begin{array}[t]{l}50.0\%\\13.71\end{array} & \begin{array}[t]{l}37.0\%\\53.51\end{array} & \begin{array}[t]{l}71.0\%\\1.44\end{array} & \begin{array}[t]{l}50.0\%\\1.99\end{array} \\ \hline
\begin{array}[t]{l}\texttt{gpt-5}\\\texttt{minimal}\end{array} & \begin{array}[t]{l}46.0\%\\12.76\end{array} & \begin{array}[t]{l}62.0\%\\1.01\end{array} & \begin{array}[t]{l}64.0\%\\7.07\end{array} & \begin{array}[t]{l}55.0\%\\14.70\end{array} & \begin{array}[t]{l}44.0\%\\2.81\end{array} & \begin{array}[t]{l}54.0\%\\14.97\end{array} & \begin{array}[t]{l}35.0\%\\53.25\end{array} & \begin{array}[t]{l}66.0\%\\1.63\end{array} & \begin{array}[t]{l}55.0\%\\2.52\end{array} \\ \hline
\begin{array}[t]{l}\texttt{gpt-5}\\\texttt{nano}\\\texttt{medium}\end{array} & \begin{array}[t]{l}32.0\%\\2.17\end{array} & \begin{array}[t]{l}60.0\%\\1.11\end{array} & \begin{array}[t]{l}49.0\%\\5.07\end{array} & \begin{array}[t]{l}45.0\%\\13.39\end{array} & \begin{array}{l}44.0\%\\3.29\end{array} & \begin{array}[t]{l}47.0\%\\13.19\end{array} & \begin{array}[t]{l}25.0\%\\43.04\end{array} & \begin{array}[t]{l}58.0\%\\1.31\end{array} & \begin{array}[t]{l}43.0\%\\1.89\end{array} \\ \hline
\begin{array}[t]{l}\texttt{gpt-5}\\\texttt{nano}\\\texttt{minimal}\end{array} & \begin{array}[t]{l}32.0\%\\2.17\end{array} & \begin{array}[t]{l}57.0\%\\1.31\end{array} & \begin{array}[t]{l}55.0\%\\5.80\end{array} & \begin{array}[t]{l}42.0\%\\12.29\end{array} & \begin{array}[t]{l}47.0\%\\3.80\end{array} & \begin{array}[t]{l}38.0\%\\10.36\end{array} & \begin{array}[t]{l}22.0\%\\32.08\end{array} & \begin{array}[t]{l}65.0\%\\1.44\end{array} & \begin{array}[t]{l}56.0\%\\2.81\end{array} \\ \hline
\begin{array}[t]{l}\texttt{gpt-5.1}\\\texttt{medium}\end{array} & \begin{array}[t]{l}46.0\%\\2.36\end{array} & \begin{array}[t]{l}65.0\%\\1.11\end{array} & \begin{array}[t]{l}55.0\%\\6.38\end{array} & \begin{array}[t]{l}52.0\%\\14.47\end{array} & \begin{array}[t]{l}49.0\%\\2.93\end{array} & \begin{array}[t]{l}52.0\%\\14.57\end{array} & \begin{array}[t]{l}35.0\%\\53.27\end{array} & \begin{array}[t]{l}65.0\%\\1.55\end{array} & \begin{array}[t]{l}51.0\%\\2.36\end{array} \\ \hline
\begin{array}[t]{l}\texttt{gpt-5.1}\\\texttt{none}\end{array} & \begin{array}[t]{l}46.0\%\\2.34\end{array} & \begin{array}[t]{l}63.0\%\\1.01\end{array} & \begin{array}[t]{l}60.0\%\\6.60\end{array} & \begin{array}[t]{l}44.0\%\\11.89\end{array} & \begin{array}[t]{l}53.0\%\\3.18\end{array} & \begin{array}[t]{l}53.0\%\\14.23\end{array} & \begin{array}[t]{l}37.0\%\\53.78\end{array} & \begin{array}[t]{l}66.0\%\\1.67\end{array} & \begin{array}[t]{l}44.0\%\\2.03\end{array} \\ \hline
\end{array}
$}
}
\end{table}

\section{Discussions}

After qualitative examination of the generated graphs, we observed that they can represent distinct functional roles across different task domains. In the SCALE MultiChallenge and AdvancedIF benchmarks, the Mermaid diagrams operate as ``procedural scaffolds,'' strictly encoding logic paths and constraint satisfaction to prevent reasoning drift. This contrasts with the GSM-Hard benchmark, where the application of our numerical masking protocol transforms the diagram into an abstract ``computational template.'' By dissociating the specific values from the structure, BRAID effectively separates the \textit{planning} of the arithmetic algorithm from its \textit{execution}. This demonstrates that the efficiency gains observed are not artifacts of answer retrieval, but rather the result of offloading the high-level cognitive architecture to the graph, allowing the solver to focus purely on atomic computation or semantic generation.

To validate that observed efficiency gains are not artifacts of memorization, we intentionally selected benchmarks with varying risks of data leakage. The GSM-Hard \cite{gao2022palr} provides essential historical baselines and can remain susceptible to inclusion in the massive web-crawled corpora used to train modern LLMs. In contrast, SCALE MultiChallenge \cite{sirdeshmukh2025multichallenge} and the AdvancedIF dataset \cite{advancedif2025} are  brand-new releases, rendering it very less likely for current pre-trained models to have encountered its solutions during training. Our proposed approach's ability to maintain high accuracy on this 'unseen' dataset provides strong empirical evidence that BRAID successfully elicits novel reasoning paths rather than relying on latent knowledge retrieval.

The most significant finding of this study as we call the BRAID Parity Effect: the observation that a smaller model equipped with bounded reasoning (BRAID) often matches or exceeds the performance of a model one or two tiers larger using unstructured prompting. For instance, on the SCALE MultiChallenge benchmark, the Nano-Medium (BRAID) outperformed the Medium (Classic) by \textit{30.31x} in PPD. This challenges the prevailing assumption that reasoning capability is strictly a function of parameter count. Instead, it suggests that reasoning performance is a product of Model Capacity × Prompt Structure. By increasing the structure, we can decrease the required capacity, democratizing access to high-quality inference.

For autonomous agents that run continuously, the PPD metrics derived in this study advocate for a split-architecture approach. Although a mermaid diagram can be handcrafted, a high-intelligence model (e.g., \texttt{gpt-4o} or \texttt{gpt-5-medium}) can be used only once to generate the BRAID graph ($C_{BRAID}$), which can then be cached. A low-cost, high-speed model (e.g., \texttt{gpt-5-nano}) can then execute this graph repeatedly ($C_{solve-only}$). Our data shows this configuration can yield efficiency gains of 30x on procedural tasks and up to 74x on mathematical reasoning compared to monolithic deployments.

\section{Future Work}
\label{sec:future_work}

While this study establishes the economic and accuracy baselines for BRAID, several avenues for structural optimization remain.

\textbf{Specialized ``Architect'' Models:} Our current approach utilizes general-purpose models (e.g., \texttt{gpt-5-medium}) to generate reasoning graphs. Future work will explore fine-tuning smaller, specialized models solely on the task of converting natural language queries into Mermaid topology. We hypothesize that a fine-tuned "Architect" model could produce higher-fidelity reasoning structures at a fraction of the current generation cost.

\textbf{Dynamic Re-planning and Self-Correction:} The current BRAID implementation treats the reasoning graph as a static artifact. We intend to investigate dynamic execution loops where the Solver model can signal a ``Topology Error'' (e.g., if no edge condition is met) and trigger a localized re-generation of the graph. This would enable agents to adapt to unforeseen variables without restarting the entire inference chain.

\textbf{Visual Graph Ingestion:} With the rise of Vision Language Models (VLMs), we plan to evaluate the efficacy of feeding the \textit{rendered} pixel-representation of the Mermaid diagram to the solver, rather than the raw code. This could leverage the strong spatial reasoning capabilities of next-generation multi-modal models.

\section{Conclusions}

This paper presented BRAID, a framework for Bounded Reasoning for Autonomous Inference and Decisions, and evaluated it across SCALE MultiChallenge, GSM-Hard, and AdvancedIF benchmarks. We demonstrated that encoding reasoning steps into structured Mermaid diagrams fundamentally alters the economics of LLM inference.

Our results show that BRAID enables "Nano" and "Mini" class models to achieve reasoning accuracy previously reserved for higher-tier models. Specifically, we observed major PPD gains on complex reasoning benchmarks like SCALE MultiChallenge and AdvancedIF datasets. We conclude that structured prompting is not merely a prompt engineering trick, but a scalable methodology for deploying reliable, cost-efficient autonomous agents. 


\subsubsection*{Acknowledgments}
This research was supported and conducted under OpenServ Labs. We thank the OpenServ team for providing computational infrastructure and API access. We thank members of Coyotiv GmbH for helpful technical discussions and feedback on the experimental design and manuscript. Additionally, we gratefully acknowledge Neol.ai for testing the BRAID framework in industrial settings and providing valuable feedback on real-world deployment.

\bibliography{bib_OpenServ}
\bibliographystyle{iclr2025_conference}

\appendix
\section{Appendix}

\subsection{BRAID Generation Prompt}
\label{app:braid_prompt}

The following system prompt is utilized in the first stage of the BRAID framework to generate the symbolic reasoning graph. The variable \texttt{\$\{conversationText\}} is replaced by the specific conversation history during inference.

\begin{center}
\begin{tcolorbox}[colback=gray!10!white, colframe=black!75!black, title=BRAID Generation Prompt]
\ttfamily \small
You are an expert at generating clear, structured Mermaid flowcharts to plan responses in multi-turn conversations.

\textbf{Task:}
\begin{itemize}
    \item Read the entire conversation history.
    \item Extract constraints, user-provided facts, references (including version references), and goals.
    \item Produce a flowchart plan that guides producing the best final assistant reply to the last user turn.
    \item Do NOT include the response itself—only the plan.
    \item Start exactly with 'flowchart TD;'
\end{itemize}

\textbf{Conversation:}\\
\$\{conversationText\}

\textbf{Output Requirements:}
\begin{enumerate}
    \item Output ONLY Mermaid code, no extra text/markdown.
    \item Start exactly with 'flowchart TD;'
    \item Each node should represent constraints, facts, or steps to produce the final reply.
    \item End nodes should indicate checks against constraints or rubric-related requirements (if implied).
\end{enumerate}
\end{tcolorbox}
\end{center}

\subsection{Performance per Dollar Analysis on SCALE MultiChallenge Dataset}

\begin{figure}[!t]
    \centering
    \includegraphics[width=\textwidth, height=0.95\textheight, keepaspectratio]{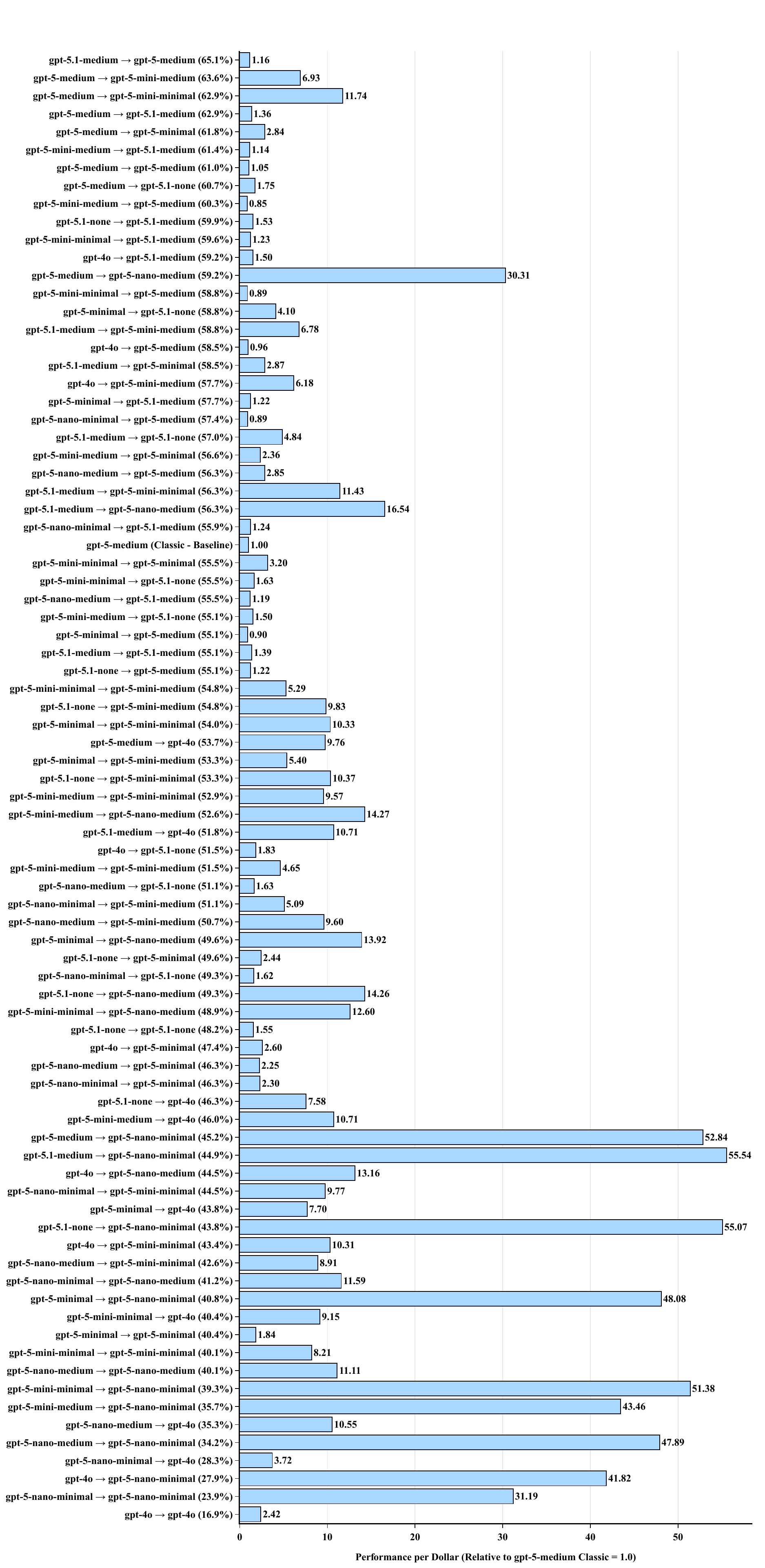}
    \caption{Performance per Dollar (PPD) for BRAID Generation $\times$ Solving combinations on the SCALE MultiChallenge dataset for \textbf{solving only} model costs. Higher values indicate major cost efficiency relative to the \texttt{gpt-5-medium} classic baseline.}
    \label{fig:ppd_scale}
\end{figure}

Figure \ref{fig:ppd_scale} illustrates the Performance per Dollar (PPD) metric for various BRAID generation and solving combinations. The PPD values include solving only cost. The values are normalized such that the baseline \texttt{gpt-5-medium (Classic)} represents a PPD of 1.00.

\textbf{High-Efficiency Pairs:} The analysis reveals that pairing a capable generator (such as \texttt{gpt-5.1-medium}) with a highly cost-effective solver (such as \texttt{gpt-5-nano-minimal}) yields the most significant economic advantage. Notably, the combination \texttt{gpt-5.1-medium} $\rightarrow$ \texttt{gpt-5-nano-minimal} achieves a PPD score exceeding 55.0, suggesting it is over 55 times more cost-effective than the baseline for this specific task.
    
\textbf{Impact of Solver Size:} There is a clear trend where smaller solvers (nano/mini variants) drastically improve the PPD ratio. This indicates that for the SCALE MultiChallenge dataset, the verification or solving step does not require the same computational overhead as the generation step to maintain acceptable accuracy.
    
\textbf{Baseline Comparison:} Homogeneous combinations using older or larger models for both generation and solving (e.g., \texttt{gpt-4o} $\rightarrow$ \texttt{gpt-4o}) generally result in lower PPD scores (often $< 3.0$), highlighting the inefficiency of using large-scale models for the solving phase in this pipeline.

\subsection{Performance per Dollar Analysis on AdvancedIF Dataset}

\begin{figure}
    \centering
     \includegraphics[width=\textwidth, height=0.95\textheight, keepaspectratio]{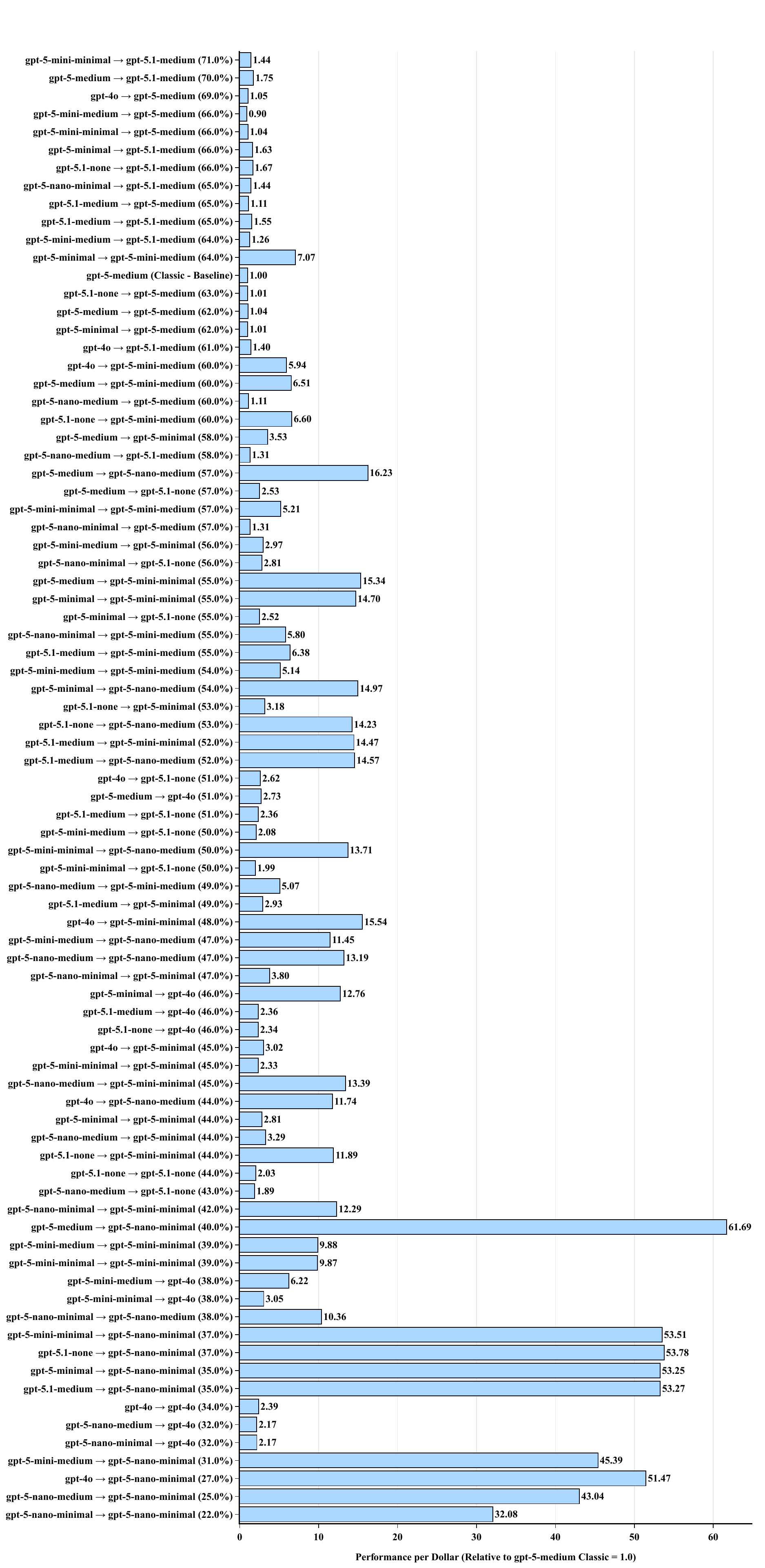}
    \caption{Performance per Dollar (PPD) for BRAID Generation $\times$ Solving combinations on the AdvancedIF dataset for \textbf{solving only} model costs. The metric highlights the  cost-efficiency of nano-scale models for this task compared to the \texttt{gpt-5-medium} baseline.}
    \label{fig:ppd_advancedif}
\end{figure}

Figure \ref{fig:ppd_advancedif} presents the Performance per Dollar (PPD) analysis for the AdvancedIF dataset. Similar to the SCALE MultiChallenge analysis, values are normalized against the \texttt{gpt-5-medium (Classic - Baseline)}.

\textbf{Dominance of Hybrid Architectures:} The data demonstrate that the highest efficiency is achieved by pairing capable generators with lightweight solvers. The combination of \texttt{gpt-5-medium} for generation and \texttt{gpt-5-nano-minimal} for solving yields the highest theoretical efficiency, with a PPD factor of \textbf{61.69}. In contrast, while the homogenous \texttt{gpt-5-nano-minimal} $\rightarrow$ \texttt{gpt-5-nano-minimal} pipeline is computationally cheap, its lower accuracy (22.0\%) limits its PPD to 32.08. This underscores that in amortized workflows, the quality of the reasoning structure (derived from a larger model) is the primary multiplier for economic efficiency.
    
\textbf{Operational ``Sweet Spot'':} A standout configuration for production reliability is the pairing of a \texttt{gpt-5-medium} generator with a \texttt{gpt-5-nano-medium} solver. This combination achieves 57.0\% accuracy—nearly matching the 63.0\% baseline—while delivering a PPD of 16.23. By utilizing the slightly more capable \texttt{gpt-5-nano-medium} rather than the \texttt{gpt-5-nano-minimal}, the system captures a 17\% absolute accuracy gain (over the 40\% achieved by \texttt{gpt-5-nano-minimal}). While this sacrifices the peak theoretical efficiency (dropping from 61.69 to 16.23 PPD), it remains over 16 times more cost-effective than the baseline. This represents a pragmatic trade-off for deployments where a minimum threshold of reasoning fidelity is required, rendering the hyper-efficient but less accurate configurations operationally impracticle.

\textbf{The Efficiency Ceiling of Large Solvers:} The analysis reveals a hard economic limit when employing large models for the solving phase. Regardless of the generator used, combinations featuring \texttt{gpt-5-medium} or \texttt{gpt-4o} as solvers consistently flatline at a PPD of approximately 1.0--2.5. This indicates that the token overhead of ingesting the structured BRAID diagram makes high-tier models economically inefficient for execution, despite their high accuracy. Consequently, the massive efficiency gains of the framework are structurally inaccessible to monolithic deployments; they can only be unlocked by transitioning the execution workload to the \texttt{nano} or \texttt{mini} tier.

\subsection{Principles of BRAID Graph Construction}
\label{sec:braid_best_practices}

The efficacy of the BRAID framework relies heavily on the quality of the underlying symbolic representation. Through our empirical evaluation on the SCALE MultiChallenge and AdvancedIF benchmarks, we identified four critical design principles for constructing effective Mermaid reasoning graphs. These practices ensure that the "Solver" model (typically a lower-parameter tier) can execute the reasoning path without Hallucination or Reasoning Drift.

\textbf{1. Node Atomicity and Token Density:}
We found that while crafting Mermaid graphs, to maximize the signal-to-noise ratio for an input prompt, each node in the graph must represent an atomic reasoning step. Traditional Chain-of-Thought often conflates observation, analysis, and conclusion into a single paragraph. In BRAID, these must be decoupled into distinct nodes (e.g., \texttt{A[Observe User Constraint]} 
→
 \texttt{B[Analyze Feasibility]}
→
 \texttt{C[Decide Strategy]}). We found that nodes containing fewer than 15 tokens yield the highest adherence rates in Nano-tier models. Verbose nodes reintroduce the noise of unstructured prompting.

\textbf{2. Procedural Scaffolding vs. Answer Leakage:}
A critical distinction exists between \textit{planning} the output and \textit{generating} the output.
\begin{itemize}
\item \textit{Ineffective (Leaking):} \texttt{Node C[Write the introduction: "Dear Team, I regret to inform you..."]}
\item \textit{Effective (Scaffolding):} \texttt{Node C[Draft Introduction: Acknowledge recent success 
→
 Pivot to financial news 
→
 Maintain regretful but professional tone]}
\end{itemize}
For creative or open-ended tasks (SCALE MultiChallenge dataset), the graph should encode the \textit{constraints} and \textit{semantic requirements} of the response, not the response text itself. An example reasoning graph is examined on the next section. This forces the Solver model to utilize its own language generation capabilities while strictly adhering to the structural bounds set by the Generator.

\textbf{3. Deterministic Branching Logic:}
Low-parameter models often struggle with ambiguity. Therefore, edges connecting nodes must be deterministic and mutually exclusive. Rather than vague transitions (e.g., \texttt{A --> B}), BRAID graphs utilize labeled edges that act as explicit condition checks (e.g., \texttt{A -- "If text > 300 words" --> B}). This transforms the inference process from a probabilistic token prediction into a directed traversal of a decision tree.

\textbf{4. Terminal Verification Loops:}
To emulate ``System 2'' thinking, effective BRAID graphs explicitly encode a ``Critic'' phase before the final output. The graph topology should converge on a set of verification nodes (e.g., \texttt{Check: Tone is empathetic}, \texttt{Check: No prohibited keywords}) before reaching the final \texttt{End} node. If a check fails, the graph should contain a feedback edge routing logic back to a revision node. This cyclic dependency enables smaller models to self-correct errors that would otherwise go unnoticed in a linear generation stream.

\subsection{Example Complexity of Generated Reasoning Graphs}
\label{app:complexity}

\begin{figure}[!t]
    \centering
     \includegraphics[width=\textwidth]{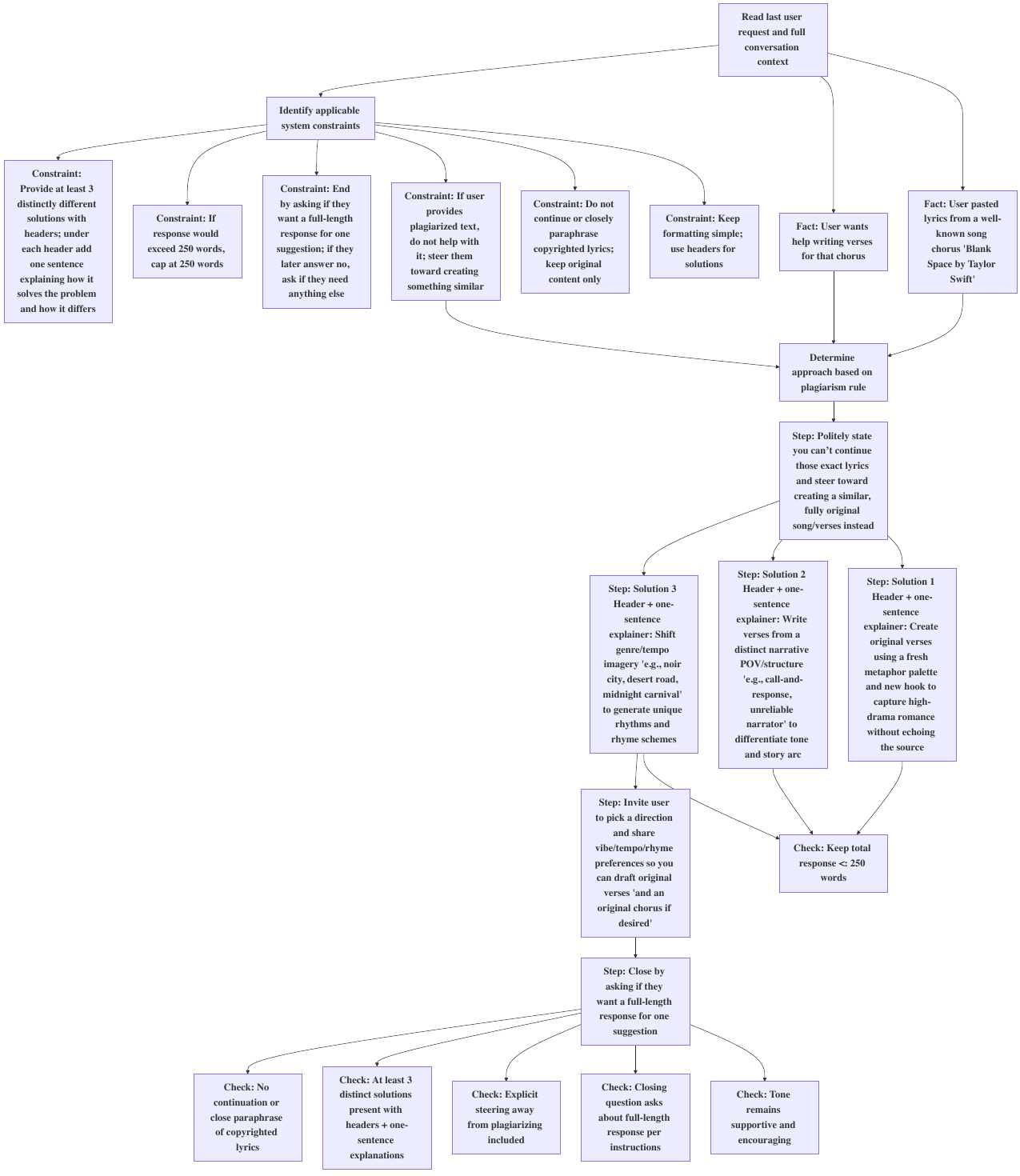}

    \caption{\textbf{BRAID-Generated Reasoning Graph for Copyright Constraint Handling for a task in AdvancedIF Dataset} This diagram depicts the automated planning process for a query involving copyrighted lyrics. The model explicitly maps out constraints, plagiarism checks, and three distinct creative solutions before executing the final response. This level of granular control is generated automatically, removing the need for human-authored decision trees.}
    
    
    \label{fig:complex_mermaid}
\end{figure}

Figure~\ref{fig:complex_mermaid} illustrates a representative reasoning graph generated by the BRAID framework for a constraint-heavy task in the AdvancedIF dataset. The task required the model to handle a request involving potentially copyrighted lyrics (Taylor Swift's ``Blank Space'') while adhering to strict formatting and plagiarism constraints.

As visualized, the resulting decision tree is highly intricate, containing over 20 distinct nodes that handle:
\begin{itemize}
    \item \textbf{Constraint Extraction:} Identifying disparate rules such as ``no plagiarism,'' ``250-word cap,'' and ``formatting requirements.''
    \item \textbf{Conditional Routing:} Branching logic that detects the user's intent to use copyrighted material and effectively steers the response toward original generation (Solutions 1, 2, and 3) rather than regurgitation.
    \item \textbf{Verification Loops:} End-nodes that serve as ``sanity checks'' for tone, length, and steering effectiveness.
\end{itemize}

Constructing a directed acyclic graph (DAG) of this complexity manually is cognitively demanding and prone to syntax errors. While manual refinement of the generated Mermaid graphs could theoretically offer further optimization, the BRAID framework simplifies the workflow by offloading the structural work to a capable generator model (e.g., \texttt{gpt-5} or \texttt{gpt-5.1}). This automation allows for the deployment of sophisticated, safe, and constraint-compliant agent behaviors that would be prohibitively expensive to hand-engineer for every unique edge case.

\clearpage
\begin{center}
\begin{tcolorbox}[colback=gray!10!white, colframe=black!75!black, title=Generated Mermaid Flowchart Code by GPT-5-Medium]
\ttfamily \small \raggedright
flowchart TD;

\vspace{0.2cm}
A[Read last user request and full conversation context] --> B[Identify applicable system constraints];\\

\vspace{0.2cm}
B --> C1[Constraint: Provide at least 3 distinctly different solutions with headers; under each header add one sentence explaining how it solves the problem and how it differs];\\
B --> C2[Constraint: If response would exceed 250 words, cap at 250 words];\\
B --> C3[Constraint: End by asking if they want a full-length response for one suggestion; if they later answer no, ask if they need anything else];\\
B --> C4[Constraint: If user provides plagiarized text, do not help with it; steer them toward creating something similar];\\
B --> C5[Constraint: Do not continue or closely paraphrase copyrighted lyrics; keep original content only];\\
B --> C6[Constraint: Keep formatting simple; use headers for solutions];\\

\vspace{0.2cm}
A --> F1[Fact: User pasted lyrics from a well-known song chorus (Blank Space by Taylor Swift)];\\
A --> F2[Fact: User wants help writing verses for that chorus];\\

\vspace{0.2cm}
F1 --> D[Determine approach based on plagiarism rule];\\
F2 --> D;\\
C4 --> D;\\

\vspace{0.2cm}
D --> S1[Step: Politely state you can't continue those exact lyrics and steer toward creating a similar, fully original song/verses instead];\\

\vspace{0.2cm}
S1 --> S2[Step: Solution 1 Header + one-sentence explainer: Create original verses using a fresh metaphor palette and new hook to capture high-drama romance without echoing the source];\\
S1 --> S3[Step: Solution 2 Header + one-sentence explainer: Write verses from a distinct narrative POV/structure (e.g., call-and-response, unreliable narrator) to differentiate tone and story arc];\\
S1 --> S4[Step: Solution 3 Header + one-sentence explainer: Shift genre/tempo imagery (e.g., noir city, desert road, midnight carnival) to generate unique rhythms and rhyme schemes];\\

\vspace{0.2cm}
S4 --> S5[Step: Invite user to pick a direction and share vibe/tempo/rhyme preferences so you can draft original verses (and an original chorus if desired)];\\
S5 --> S6[Step: Close by asking if they want a full-length response for one suggestion];\\

\vspace{0.2cm}
S2 --> WC[Check: Keep total response <= 250 words];\\
S3 --> WC;\\
S4 --> WC;\\
S6 --> D1[Check: No continuation or close paraphrase of copyrighted lyrics];\\
S6 --> D2[Check: At least 3 distinct solutions present with headers + one-sentence explanations];\\
S6 --> D3[Check: Explicit steering away from plagiarizing included];\\
S6 --> D4[Check: Closing question asks about full-length response per instructions];\\
S6 --> D5[Check: Tone remains supportive and encouraging];
\end{tcolorbox}
\end{center}

\end{document}